\pdfoutput=1

\documentclass[9.5pt,journal,compsoc,cspaper]{IEEEtran}
\usepackage[pdftex]{graphicx}
\usepackage{booktabs, makecell, tabularx}
\usepackage{rotating}
\usepackage[export]{adjustbox}
\usepackage{gensymb}
\usepackage{ragged2e}
\usepackage{paralist}
\usepackage{multirow}
\usepackage{color}
\usepackage[pagebackref=true,breaklinks=true,letterpaper=true,colorlinks,bookmarks=false]{hyperref}

\definecolor{todocolor}{RGB}{0,174,247}

\ifCLASSOPTIONcompsoc
  \usepackage[nocompress]{cite}
\else
  \usepackage{cite}
\fi
\ifCLASSINFOpdf
  \usepackage[pdftex]{graphicx}
  \graphicspath{{../pdf/}{../jpeg/}{./figures}}
\else
\fi
\usepackage{xspace}
\usepackage{pgfplots}
\usepackage{xcolor}
\usepackage{tikz}

\usepackage{amsmath}
\usepackage{amssymb}
\usepackage{array}

\ifCLASSOPTIONcompsoc
  \usepackage[caption=false,font=footnotesize,labelfont=sf,textfont=sf]{subfig}
\else
  \usepackage[caption=false,font=footnotesize]{subfig}
\fi

\ifCLASSOPTIONcaptionsoff
  \usepackage[nomarkers]{endfloat}
 \let\MYoriglatexcaption\caption
 \renewcommand{\caption}[2][\relax]{\MYoriglatexcaption[#2]{#2}}
\fi
\usepackage{url}
\usepackage{etex}

\renewcommand{\baselinestretch}{.985}

\begin{document}
\bstctlcite{IEEEexample:BSTcontrol}
%

\title{Learning to Extract Building Footprints \\from Off-Nadir Aerial Images}

%
%
%
%

\author{Jinwang~Wang,
        Lingxuan~Meng,
        Weijia~Li,
        Wen~Yang, 
        Lei~Yu, 
        Gui-Song~Xia

\IEEEcompsocitemizethanks{\IEEEcompsocthanksitem J. Wang is with the School of Electronics, Wuhan University, China and also with SenseTime Research, China. E-mail: jwwangchn@whu.edu.cn.
\IEEEcompsocthanksitem W. Yang, L. Yu are with the School of Electronics, Wuhan University, China. E-mail: \{yangwen, ly.wd\}@whu.edu.cn.
\IEEEcompsocthanksitem L. Meng is with University of Electronic Science and Technology of China, China and also with SenseTime Research, China. E-mail: xuanxuanling@std.uestc.edu.cn.
\IEEEcompsocthanksitem W. Li is with Sun Yat-Sen University and also with CUHK-SenseTime joint Lab, China. E-mail: liweij29@mail.sysu.edu.cn.
\IEEEcompsocthanksitem G. S. Xia is with the School of Computer Science, Wuhan University, China.
E-mail:guisong.xia@whu.edu.cn.
\IEEEcompsocthanksitem The studies in this paper have been supported by the NSFC projects under the contracts No.61771351, No.61771350 and No.61922065.
\IEEEcompsocthanksitem Corresponding authors: W. Yang and G. S. Xia.
}
}

%
%

\markboth{Journal of \LaTeX\ Class Files,~Vol.~XX, No.~XX, January~2021}%
{Shell \MakeLowercase{\textit{et al.}}: Bare Demo of IEEEtran.cls for Computer Society Journals}
%



\newcommand{\ie}{\textit{i.e.}\xspace}
\newcommand{\eg}{\textit{e.g.}\xspace}
\newcommand{\etal}{\textit{et al.}\xspace}
\newcommand{\wrt}{\textit{w.r.t.}\xspace}
\newcommand{\etc}{\textit{etc.}\xspace}
\newcommand{\resp}{\textit{resp.}\xspace}

\newcommand{\fixedvskip}{-3mm}
\newcommand{\fixedvskiptab}{-3mm}

\newcommand{\equspace}{6pt}
\newcommand{\revision}[1]{{\color{blue}{#1}}}
\newcommand{\smallrevision}[1]{{\color{black}{#1}}}



\definecolor{todocolor}{RGB}{0,174,247}

\IEEEtitleabstractindextext{%
\begin{abstract}
    \justifying
    Extracting building footprints from aerial images is essential for precise urban mapping with photogrammetric computer vision technologies. Existing approaches mainly assume that the roof and footprint of a building are well overlapped, which may not hold in off-nadir aerial images as there is often a big offset between them. In this paper, we propose an offset vector learning scheme, which turns the building footprint extraction problem in off-nadir images into an instance-level joint prediction problem of the building roof and its corresponding ``{\em roof to footprint}" offset vector. Thus the footprint can be estimated by translating the predicted roof mask according to the predicted offset vector. We further propose a simple but effective feature-level offset augmentation module, which can significantly refine the offset vector prediction by introducing little extra cost. Moreover, a new dataset, Buildings in Off-Nadir Aerial Images (BONAI), is created and released in this paper. It contains 268,958 building instances across 3,300 aerial images with fully annotated instance-level roof, footprint, and corresponding offset vector for each building. Experiments on the BONAI dataset demonstrate that our method achieves the state-of-the-art, outperforming other competitors by $3.37$ to $7.39$ points in F1-score. The codes, datasets, and trained models are available at \url{https://github.com/jwwangchn/BONAI.git}.
\end{abstract}

\begin{IEEEkeywords}
Building footprint extraction, building detection, learning offset vector, off-nadir aerial image.
\end{IEEEkeywords}}

\maketitle

\IEEEdisplaynontitleabstractindextext

%
\IEEEpeerreviewmaketitle


\section{Introduction}
\label{sec:introduction}

\IEEEPARstart{A}{utomatic} Building Footprint Extraction (BFE) from aerial images for urban scenes has been studied for decades and benefited a wide variety of geomatics and Earth observation tasks, \eg, 3D city modeling, building change detection and precise urban planning~\cite{Birth_Death-2011-TPAMI, SDF_2020_CVPR,Signed_distance_function_2017_TPAMI, DSAC_CVPR_2018, Conv_MPN_CVPR_2020}.

\begin{figure}[t!]
	\centering
	\includegraphics[width=1.0\linewidth]{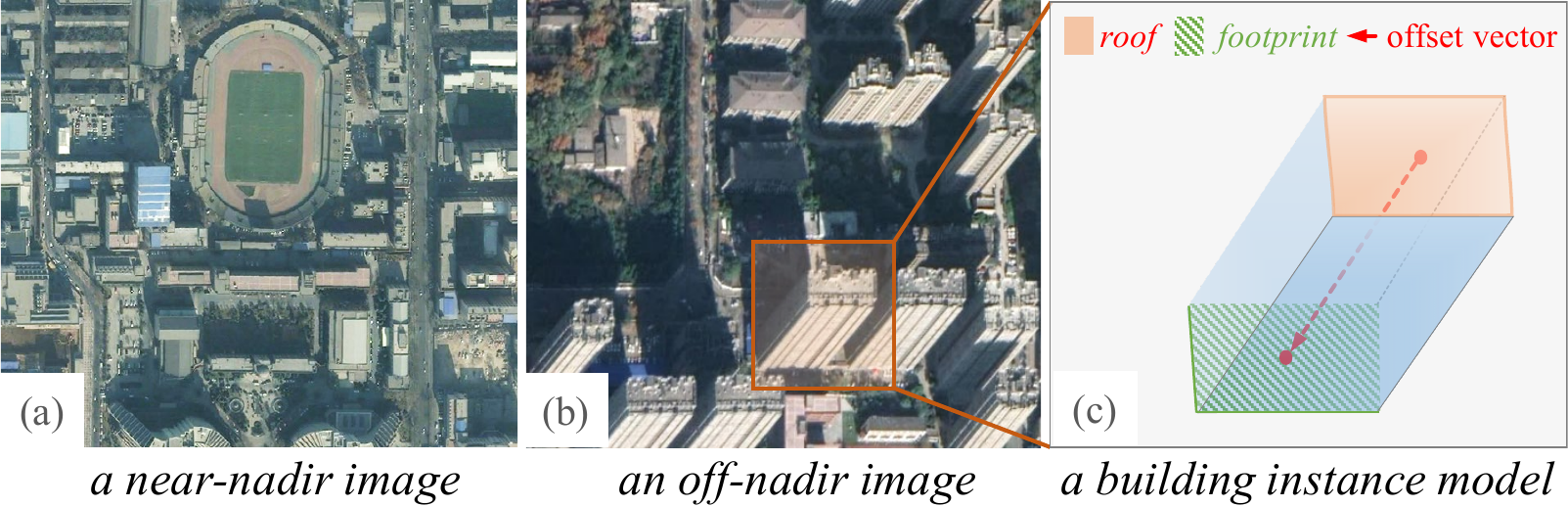}
    \vspace{-5mm}
	\caption{An illustration of the BFE problem in off-nadir images. In contrast with those in near-nadir images (a), for a building in off-nadir images (b), 
	there are often non-negligible offsets between the projections of the roof and the footprint, and the boundary of the building footprint is usually partially visible due to the heavy  occlusion by its facade. In this paper, we present a building instance model (c) by relating the occluded footprint to the visible roof of the building with a learned offset vector.
 }
	\label{fig:near_and_off_nadir}
    \vspace{-4mm}
\end{figure}

Early work tackles the BFE problem primarily through exploiting image structure and appearance features to characterize building footprints~\cite{BFE3_1999_TPAMI,BFE1_2007_TPAMI,svm_building_2007_ISPRSJ,BFE2_2008_TPAMI}, while the performance of these methods is often limited by the discriminative capability of shallow features. Recently, deep learning-based solutions~\cite{Signed_distance_function_2017_TPAMI,Darnet_2019_CVPR, MAP-Net_2020_TGRS,Polygon_Building_2020_CVPR} have reported promising results on BFE, benefiting from the powerful capability of deep models in representation learning. However, the BFE methods mentioned above mainly focus on near-nadir images, as shown in Fig.~\ref{fig:near_and_off_nadir} (a), where the projected positions of the roof and the footprint of a building are usually well overlapped, and the BFE problem boils down to the extraction of the visible roofs of buildings. Few of them can handle off-nadir images that are often acquired when the viewing angle of the aerial imaging system is large (\eg, larger than $25^\circ$). An example of off-nadir images and a building instance model in it are illustrated in Fig.~\ref{fig:near_and_off_nadir} (b) and (c) respectively. It can be observed that, for a building in off-nadir images, there is often a non-negligible offset between the projections of the roof and the footprint. Moreover, the boundary of the building footprint is partially visible and heavily occluded by its facade. All these pose great challenges to accurately detecting footprints of buildings in off-nadir images.

Despite the difficulties, it is actually in great demand to extract pixel-level building footprints from off-nadir images, see \eg, the \textit{SpaceNet 4 Challenge}\footnote{https://spacenet.ai/off-nadir-building-detection/}. So far, to the best of our knowledge, few methods can accurately extract the obscured contour of the building footprints, except the work~\cite{Learning_Geocentric_Object_Pose_2020_CVPR} that uses off-nadir images with the help of LiDAR data to learn the geocentric pose of a building for generating its pixel-level footprint. However, compared with imagery, the LiDAR data is much less accessible. 
In this paper, we concentrate on recovering accurate building footprints from off-nadir images, 
in a more general setting that only two-dimensional images are available. 

With a preliminary observation that the contours of the roof and the footprint are often consistent for most single buildings in the urban scenes, as shown in Fig.~\ref{fig:near_and_off_nadir}, we propose to learn the partially occluded footprints of buildings from their visible roofs and facades. More precisely, instead of directly extracting building footprints, we tackle the BFE problem in off-nadir images by simultaneously learning an instance-level building roof and its corresponding offset vector toward the footprint for each building. To do so, we present a novel model named {\em Learning OFfset vecTor} (LOFT). Specifically, we design a new {\em offset head} to predict the ``{\em roof to footprint}" offset vector for each building, which can be easily applied to typical top-down instance segmentation methods such as Mask R-CNN~\cite{Mask-R-CNN_2017_ICCV}. The building footprints are then estimated by translating the predicted roof masks according to the predicted offset vectors. 
As the roofs are visible for most buildings, and the offset vectors are actually embedded in the visible facade structures, our LOFT method can achieve better BFE performance than the methods that directly model the building footprints. 

Noticing that the length of the ``{\em roof to footprint}" offset vector mainly depends on the building height, while the angle of the offset vector is sensitive to the perspective angle of the imaging system, it may not converge well when training the offset head with limited offset training samples. Moreover, the performance of BFE largely depends on the accuracy of the offset prediction in the LOFT scheme. Therefore, we further present a simple yet effective {\em Feature-level Offset Augmentation} (FOA) module to reduce the offset prediction errors. Unlike traditional image-based rotation augmentation, the proposed FOA module is implemented via simple rotations operating in the abstract feature space thus with very limited sacrifice of computational cost. Specifically, in the training stage, the offset vector and its corresponding offset feature of each building are synchronously rotated by multiple angles, based on which multiple offset losses are calculated to train an Offset Prediction Network (OPN). In the inference stage, the multiple offsets predicted by the OPN are fused to output a final offset.

Currently, there are mainly two types of datasets to train and evaluate BFE models: {\em pixel-level} labeled datasets, \eg~\cite{INRIA_2017_IGARSS,ISPRS_2018} and {\em instance-level} labeled datasets, \eg~\cite{DSTL_2018_kaggle,WHU_2018_TGRS,SpaceNet_MVOI_2019_CVPR}. Among the first type, the {\em INRIA Aerial Image Labeling dataset}~\cite{INRIA_2017_IGARSS} provides pixel-wise labels of building/non-building classes, while the {\em ISPRS Benchmark dataset}~\cite{ISPRS_2018} annotates images with six categories, including buildings. Both are prepared to train and evaluate BFE models which tackle the BFE problem through a pixel-level segmentation task. On the other hand, the {\em DSTL Kaggle Dataset}~\cite{DSTL_2018_kaggle} and the WHU Building dataset~\cite{WHU_2018_TGRS} annotate buildings in aerial images with polygons, which are suitable to train and evaluate instance segmentation-based BFE methods.
Note that, except for~\cite{SpaceNet_MVOI_2019_CVPR}~and~\cite{Synthinel-1_2020_WACV}, which have some off-nadir images, the  datasets above contain almost only near-nadir images and few off-nadir images. Moreover, those datasets are only annotated with building footprints, and the offset vectors are missed. 

To train and evaluate our LOFT method for off-nadir images, the annotations of building roofs and their corresponding offset vectors are essential. Therefore, we create a new dataset for BFE in off-nadir imagery, dubbed as BONAI (\textbf{B}uildings in \textbf{O}ff-\textbf{N}adir \textbf{A}erial \textbf{I}mages). The BONAI dataset contains 268,958 building instances across 3,300 aerial images. Unlike the datasets mentioned above, on the one hand, BONAI contains a large amount of off-nadir images. On the other hand, the images in BONAI are annotated with building roof, building footprint, and the automatically generated offset vector. Note that due to the consistent contour of roof and footprint, the annotation cost of BONAI dataset is slightly higher than those who only annotate the building footprints. The detailed annotation process will be described in Sec.~\ref{sec::dataset}.

In experiments, we leverage our proposed offset head on the existing state-of-the-art instance segmentation methods (\ie, Mask R-CNN~\cite{Mask-R-CNN_2017_ICCV}, PANet~\cite{PANet_2018_CVPR}, Cascade Mask R-CNN~\cite{HTC_2019_CVPR}, HRNet-v2~\cite{HRNet_2019_arXiv}) to demonstrate that the LOFT scheme can fully exploit the structural information of buildings to obtain precise footprint contours. In the standard evaluation of building footprint extractors, our proposed offset head and FOA consistently improve the F1-Score of the aforementioned extractors on BONAI dataset. Furthermore, we evaluate the LOFT scheme on the near-nadir BFE dataset to verify its generalization.

To summarize, our main contributions are three-fold:
\begin{compactitem}
	\setlength\itemsep{1pt}
    \item We propose to cast the BFE problem in off-nadir images as a problem of estimating the instance-level building roof and predicting its offset vector simultaneously, and present a new model, \ie, LOFT, which can be used for both off-nadir and near-nadir images.
    \item We further present a simple yet effective feature-level offset augmentation module to refine the offset vector prediction through transforming the input features in the abstract feature space, which only requires slightly incremental computation.
    \item We introduce a new well-annotated dataset for BFE in off-nadir imagery, \ie, BONAI, in which buildings are well annotated with instance-level roofs, footprints and corresponding offset vectors.
\end{compactitem}

\section{Methodology}
\label{sec::methodology}

After giving a general setup of the BFE problem in off-nadir images by Sec.~\ref{sec:problem_definition}, this section first presents 
an overview of our method in Sec.~\ref{sec:overview}. All details of the proposed method are subsequently described in Sec.~\ref{sec:SV-LOVE} and Sec.~\ref{sec:foa}.

\subsection{Problem Setup}
\label{sec:problem_definition}
Given an off-nadir aerial image $\mathbf I$, the task of BFE is to design a model to locate the footprints and simultaneously extract their boundaries of all buildings contained by $\mathbf I$. As mentioned before, the main difficulty lies in the fact that the footprints of buildings are often partially visible when imaging at off-nadir viewing angles.

In this work, we propose to solve this problem by supervised learning a deep BFE model with a set of $N$ instance-level labeled data ${\mathcal{D}}$, {\em i.e.}, 
\begin{equation}
	\setlength{\abovedisplayskip}{\equspace}
	\setlength{\belowdisplayskip}{\equspace}
    {\mathcal{D}} = \Big\{(\mathbf{I}_i, \mathcal{T}_i); \, i =1, \ldots, N \Big\},
\end{equation} 
where $\mathcal{T}_i$ is the corresponding label of buildings contained by the image $\mathbf{I}_i$ in ${\mathcal{D}}$. In particular, $\mathcal{T}_i$ consists of $K$ footprints $\{{\bf f}_i^j; \, j =1, \ldots, K \}$ with each ${\bf f}_i^j \in \mathbb{R}^{2\times M}$ being represented by a polygon of $M$ vertexes.

Moreover, as the footprints of buildings are partially visible while the roof is always fully visible, the problem can be converted into estimating the roof associated with an offset vector to the footprint for each building. Thus, every footprint label ${\bf f}_i^j $ in the dataset ${\mathcal{D}}$ corresponds to a roof label ${\bf r}_i^j $ and an extra offset label ${\bf o}_i^j\in \mathbb{R}^2$, which represents the offset vector from the roof to the footprint of the $j$-th building in the image $\mathbf{I}_i$. Therefore, the labeled data will be,
\begin{equation}
	\setlength{\abovedisplayskip}{\equspace}
	\setlength{\belowdisplayskip}{\equspace}
    \mathcal{D} = \Big\{\big(\mathbf{I}_i, \{{\bf f}_i^j, \, {\bf r}_i^j, \, {\bf o}_i^j\}_{j=1}^K \big); \, i =1, \ldots, N \Big\}.
\end{equation}
An example of the instance-level annotation of a building in off-nadir images is demonstrated in Fig.~\ref{fig:near_and_off_nadir} (c).

\subsection{Overview}
\label{sec:overview}

\begin{figure}[t!]
	\centering
	\includegraphics[width=.92\linewidth]{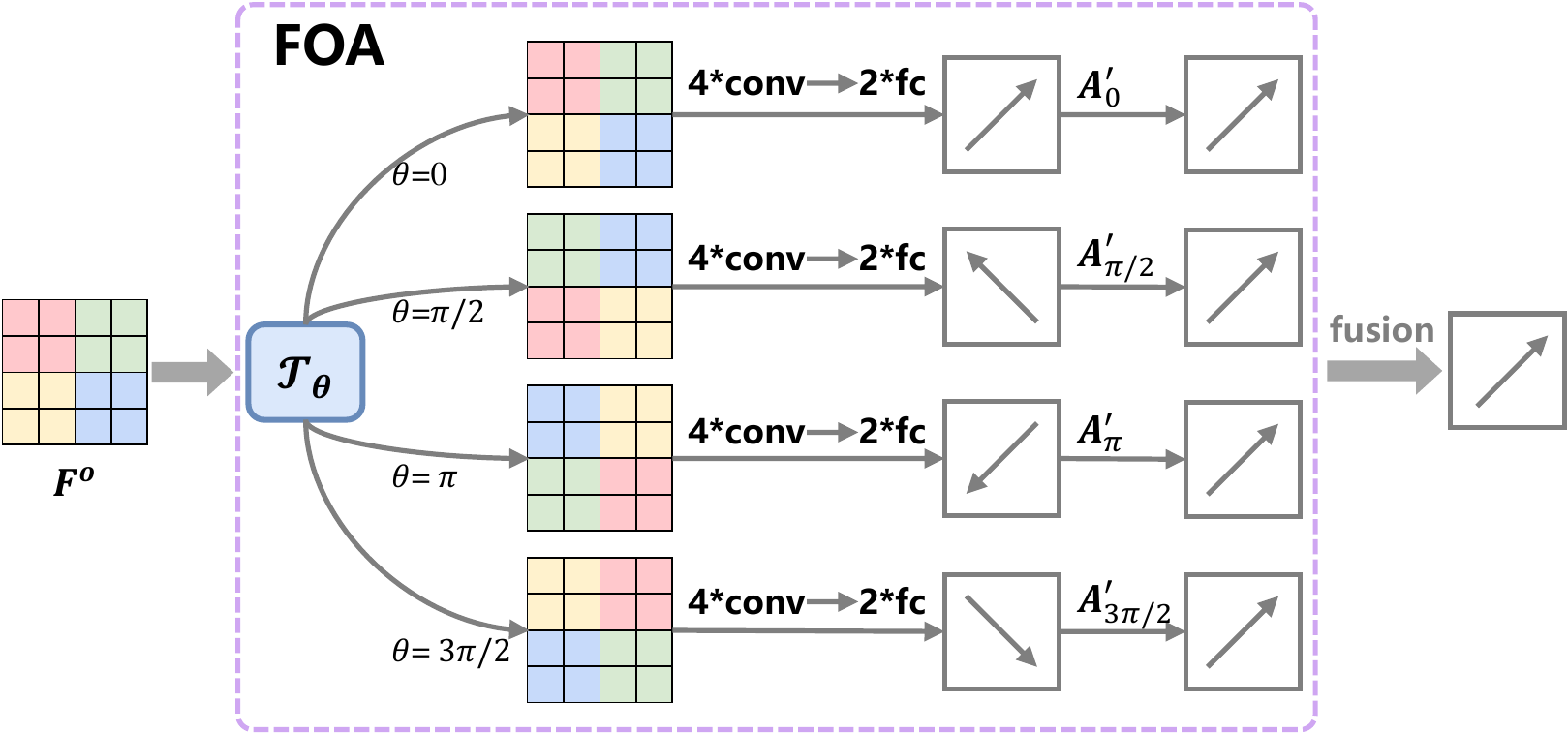}
    \vspace{-4mm}
	\caption{The architecture of the FOA module. Conv and FC denote the convolution (Conv) layer and fully connected (FC) layer, respectively. There are four branches, and the parameters of FC layers are shared. In the training stage, each branch rotates the input feature map $\mathbf{F}^{o}$ by the given rotation angle and regresses the corresponding offset vector. In the inference stage, the offset vectors regressed by four branches are rotated inversely and fused to form the final offset vector.}
	\label{fig:FOA}
 	\vspace{-3mm}
\end{figure}

We elaborate our LOFT scheme by casting the BFE problem in off-nadir images as a problem of estimating the instance-level building roof and predicting its offset vector simultaneously. To learn the instance-level offset vector of each building, LOFT introduces an {\em offset head} to the top-down instance segmentation framework as shown in Fig.~\ref{fig:SV-LOVE-framework}. To refine the offset vector prediction for mitigating the effects of noises raised from the image acquisition and offset learning, we further propose an efficient feature-level offset augmentation module, displayed in Fig.~\ref{fig:FOA}, by feature transformations in the offset head.

\subsection{Learning Offset Vector (LOFT)}
\label{sec:SV-LOVE}

The LOFT model approaches the BFE problem in off-nadir images by predicting the roofs and their associated offset vectors to the footprints of buildings, with the supervision of dataset ${\mathcal{D}}$. The idea is conceptually simple: plugging a new {\em offset head} to learn the offset vector into a top-down instance segmentation framework used for learning the building roofs. In what follows, the Mask R-CNN~\cite{Mask-R-CNN_2017_ICCV} is adopted to demonstrate the proposed LOFT model for its simplicity. Similarly, our proposed LOFT module can also be used in some other instance segmentation frameworks.

\begin{figure}[t!]
\centering
	\includegraphics[width=.95\linewidth]{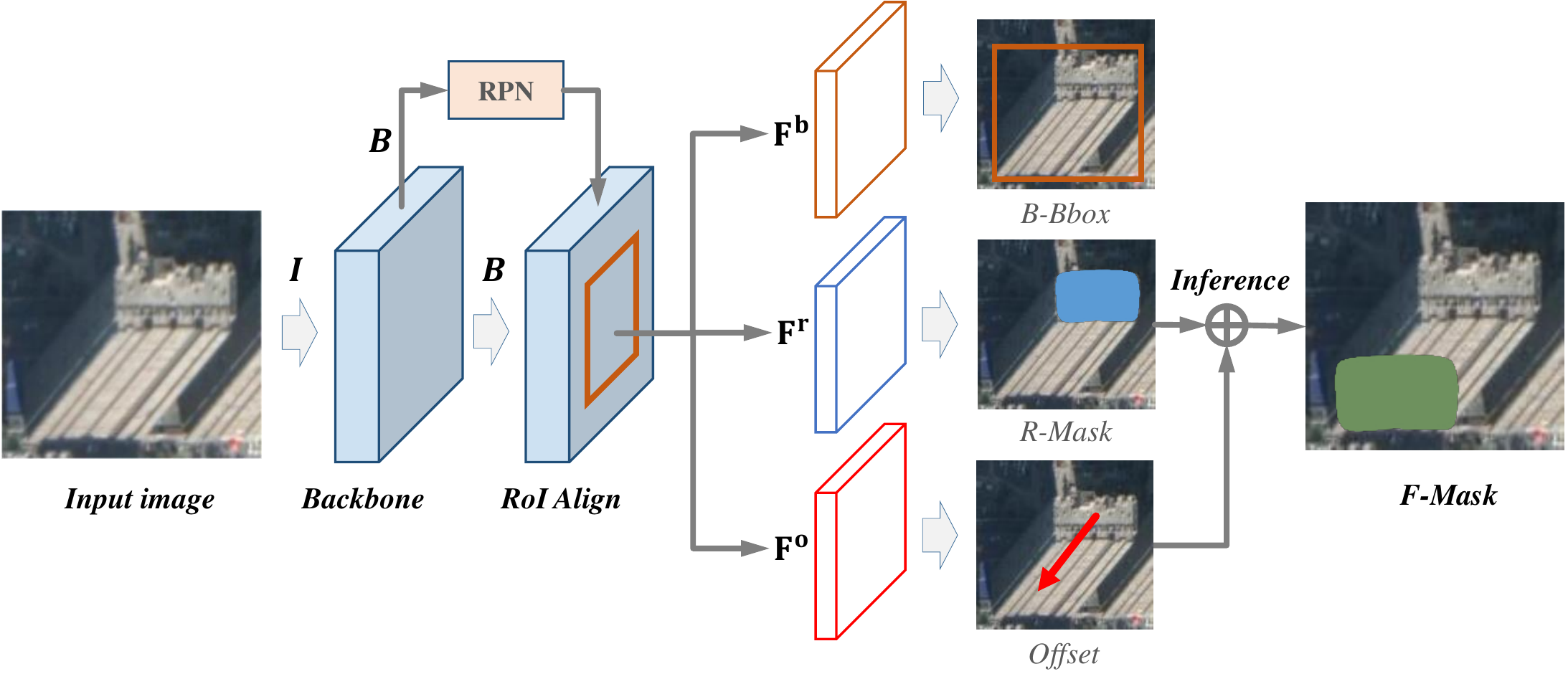}
  	\vspace{-4mm}
	\caption{The architecture of the proposed LOFT scheme (based on Mask R-CNN). The backbone is the ResNet with Feature Pyramid Network (FPN), and RPN stands for Region Proposal Network. The RoI Align layers produce the feature maps $\mathbf{F}^{b}$, $\mathbf{F}^r$, and $\mathbf{F}^{o}$ to generate the building bounding box (B-Bbox), the roof mask (R-Mask), and the offset vector (Offset). The footprint mask (F-Mask) is finally computed with the predicted R-Mask and Offset in the inference stage.}
	\label{fig:SV-LOVE-framework}
 	\vspace{-3mm}
\end{figure}

The overall architecture of LOFT is illustrated in Fig.~\ref{fig:SV-LOVE-framework}. To train the LOFT model, the ground-truth labels of building bounding boxes (B-Bbox) $\{{\bf b}_i^j\}_{j=1}^K$, the roof masks (R-Mask) $\{{\bf r}_i^j\}_{j=1}^K$, and the offset vectors (Offset) $\{{\bf o}_i^j\}_{j=1}^K$ are required, where each ${\bf b}_i^j\in \mathbb{R}^4$ is a bounding box denoted by $\{(x_i^j, y_i^j), w_i^j, h_i^j\}$ and each ${\bf r}_i^j \in \mathbb{R}^{2\times M}$ is represented by a polygon of $M$ vertexes. This information can be obtained from the labels $\mathcal{T}_i = \{{\bf f}_i^j, \, {\bf o}_i^j\}_{j=1}^K$. The detailed generation process is described in Sec.~\ref{sec::dataset}.

In the training stage, with a set of labeled data ${\mathcal{D}}$, an input image $\mathbf{I}$ is fed into the backbone network, which produces the backbone feature map $\mathbf{B}$ as shown in Fig.~\ref{fig:SV-LOVE-framework}. Then, the Region Proposal Network (RPN)~\cite{Faster-R-CNN_2015_NIPS} takes the feature map $\mathbf{B}$ as the input to generate region proposals $\mathcal{P}$. Next, three RoI Align~\cite{Mask-R-CNN_2017_ICCV} layers take each proposal $p_i=(x_i, y_i, w_i, h_i)\in \mathcal{P}$ and $\mathbf{B}$ as inputs to compute feature maps $\mathbf{F}^{b}$, $\mathbf{F}^{r}$, and $\mathbf{F}^o$, where $p_i$ denotes the $i$-th building proposal, $\mathbf{F}^{b}$, $\mathbf{F}^{r}$, and $\mathbf{F}^o$ are the building bounding box, roof mask, and offset vector feature maps in the R-CNN head, mask head, and {offset head}, respectively. The R-CNN and mask heads are inherited from the Mask R-CNN~\cite{Mask-R-CNN_2017_ICCV}. For the {offset head}, the feature map $\mathbf{F}^o$ is used as the input, and it consists of several convolution (Conv) layers and fully connected (FC) layers. Note that the ground truths of the RPN and the R-CNN head are building bounding boxes, since the receptive field of the network in the {offset head} needs to cover roof and footprint at the same time for regressing the offset vector, \ie, the information of offset vector is embedded in the visible building facade structure.

The LOFT model is finally obtained via minimizing a joint loss function,
\begin{equation}
	\setlength{\abovedisplayskip}{\equspace}
	\setlength{\belowdisplayskip}{\equspace}
	\mathcal{L} = \mathcal{L}_{\textrm{RPN}} + \alpha_1\mathcal{L}_{\textrm{R-CNN}} + \alpha_2\mathcal{L}_{\textrm{Mask}} + \alpha_3\mathcal{L}_{\textrm{Offset}},
	\label{equ:offset_rcnn_joint_loss}
\end{equation}
where $\mathcal{L}_{\textrm{RPN}}$, $\mathcal{L}_{\textrm{R-CNN}}$, $\mathcal{L}_{\textrm{Mask}}$ are the same as those in Mask R-CNN, \ie, the losses for the RPN, R-CNN, and mask heads, respectively. $\mathcal{L}_{\textrm{Offset}}$ is the loss for the offset head, {where a standard} smooth $L_1$ Loss is used. We empirically set the loss weights as $\alpha_1=1, \alpha_2=1$, and $\alpha_3=2$ in the experiments. 

To speed up the {offset head} convergence, the following encoding functions are used:
\begin{equation}
	\setlength{\abovedisplayskip}{\equspace}
	\setlength{\belowdisplayskip}{\equspace}
	\phi_{x} = o_x/w^{\textrm{p}},\quad
	\phi_{y} = o_y/h^{\textrm{p}},
\end{equation}
where $w^p$ and $h^p$ are the width and height of the matched proposal $p_m\in\mathcal{P}$, $[o_x, o_y]^T$ is the ground truth offset vector, and $[\phi_x, \phi_y]^T$ is the encoded offset vector for regression.

During the inference stage, we use the predicted offset vectors to convert the predicted roof masks to the footprint masks. Specifically, a predicted roof mask will firstly be represented as a polygon $ \mathbf r \in \mathbb{R}^{2 \times M}$ with $M$ vertexes by a topological structural analysis algorithm~\cite{Topological_Structural_Analysis_1985_CVGIP}. The footprint polygon $\mathbf f$ is finally computed by translating the roof polygon $\mathbf{r}$ with the predicted offset $\hat{o}=[\hat{o}_x, \hat{o}_y]^T$.

\subsection{Feature-level Offset Augmentation (FOA)}
\label{sec:foa}

Note that the offset vector $\mathbf o=[o_x, o_y]^T$ can be converted into $[o_{\rho}, o_{\theta}]^T$ in a polar coordinate system. The $o_{\theta}$ is approximately uniformly distributed since the perspectives of the aerial imaging platforms are almost arbitrary to the scenes. Hence, the network needs to handle arbitrary rotation transformations when learning offsets. Moreover, the {offset head} may not well converge with a limited number of offset training samples. Thus designing an enhancement module is necessary to learn more robust offset features. One way is 
using image-level rotation augmentations. Observing that the offset vectors in an image can only be rotated by one angle in a training epoch and the network thus needs more time to converge, we present an FOA module by rotating the offset features, improving the robustness of offset prediction by operating in the abstract feature space as in ~\cite{Featmatch_2020_ECCV} .

The architecture of the FOA module is shown in Fig.~\ref{fig:FOA}. We extend the single forward {offset head} to multiple parallel {offset head} branches and these branches regress multiple rotated offset vectors by different angles in parallel. Specifically, the input feature map $\mathbf{F}^{o}$ and corresponding ground-truth offset vector $[o_x, o_y]^T$ will be simultaneously rotated by a rotation angle set $\Theta=\{\theta_1, \theta_2, \cdots, \theta_n\}$. 

Specifically, we exploit the spatial transformer~\cite{STN_2015_NIPS} for the feature map rotation. Firstly, the rotation angle $\theta_i \in \Theta$ is used to create a sampling grid, a set of points where the input feature map is sampled to produce the transformed feature map.
More precisely, given the rotation matrix,
\begin{equation}
	\setlength{\abovedisplayskip}{\equspace}
	\setlength{\belowdisplayskip}{\equspace}
	A_\theta = \left[
	\begin{array}{crc}
	\cos \theta & -\sin \theta\\
	\sin \theta & \cos \theta
	\end{array}
	\right],
	\label{equ:rotation_matrix}
\end{equation}
the point-wise transformation on an input feature map is 
\begin{equation}
	\setlength{\abovedisplayskip}{\equspace}
	\setlength{\belowdisplayskip}{\equspace}
	\left[x_i^s, \,y_i^s \right]^T
	= A_\theta
	\left[x^t_i, \, y^t_i \right]^T,
	\label{eqn:offset_rotation}
\end{equation}
where $[x_i^t, y_i^t]^T$ is the target coordinate of the regular grid in the output feature map, and $[x_i^s, y_i^s]^T$ is the source coordinate in the input feature map.
A similar way is applied to the offset vector rotation, \ie,
a rotated offset vector $\mathbf{o}^* =[o_x^*, o_y^*]^T$ is computed by $\mathbf{o}^* = A_\theta \mathbf{o}$.

Fig.~\ref{fig:FOA} illustrates the network architecture of the FOA module with a rotation angle set $\{0, \frac{\pi}{2}, \pi, \frac{3\pi}{2}\}$, each branch of which consists of a series of Conv and FC layers with the same parameters as the {offset head} in Sec.~\ref{sec:SV-LOVE}. To reduce the parameters in the FOA module, the parameters of FC layers in all branches are shared. In the training stage, each branch will calculate offset vector loss separately, while in the inference stage, the predicted multiple offset vectors in corresponding branches will rotate reversely according to the rotation angle. The final fused offset is generated by the {\em max} selection strategy since we find that the values of offsets tend to be smaller than ground-truth values.
\section{Experiments and Analysis}
\label{sec:experiments}

\begin{figure}[t!]
    \centering
    \newcommand{\rate}{0.49}
    \renewcommand{\tabcolsep}{0.5mm}
    \renewcommand{\arraystretch}{0.6}
    \centering
	\begin{tabular}{cc}
    	\includegraphics[width=\rate\linewidth]{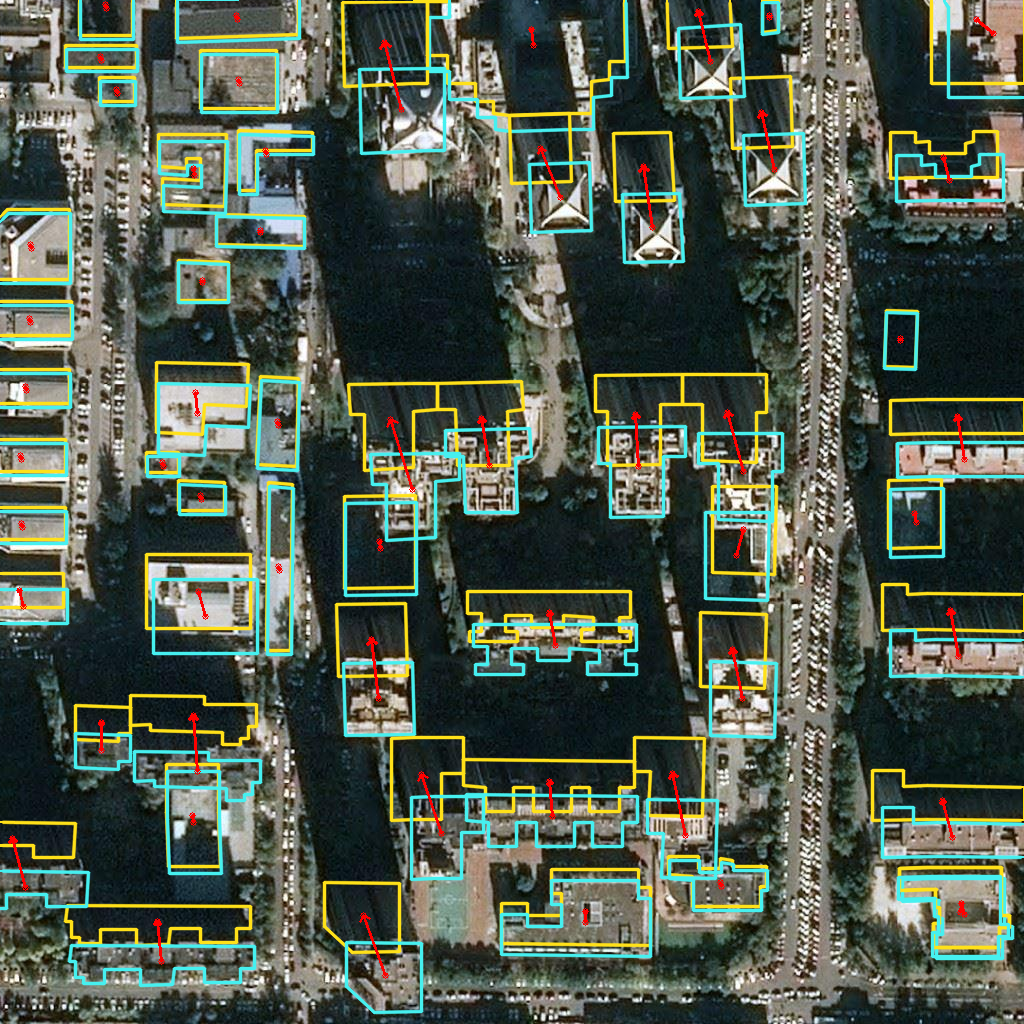} &
        \includegraphics[width=\rate\linewidth]{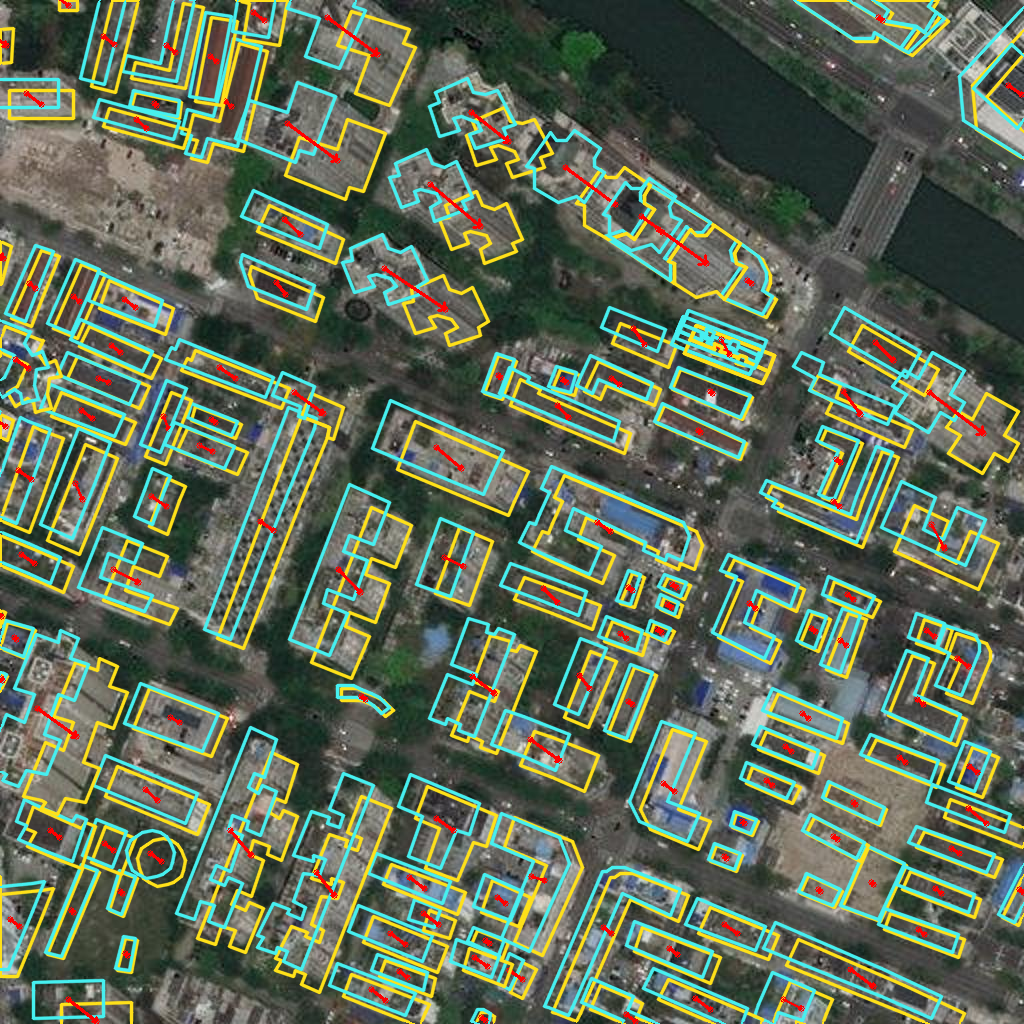}
        \\
        \includegraphics[width=\rate\linewidth]{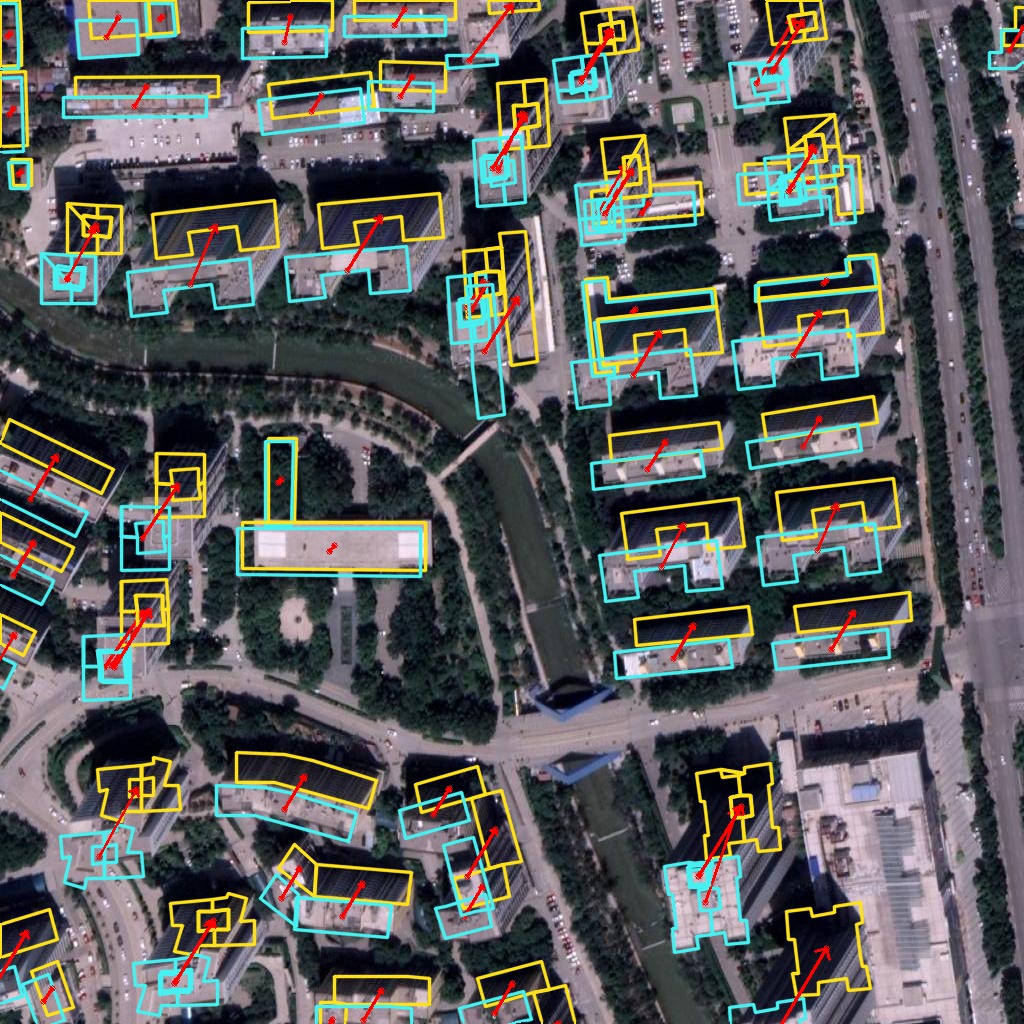} &
        \includegraphics[width=\rate\linewidth]{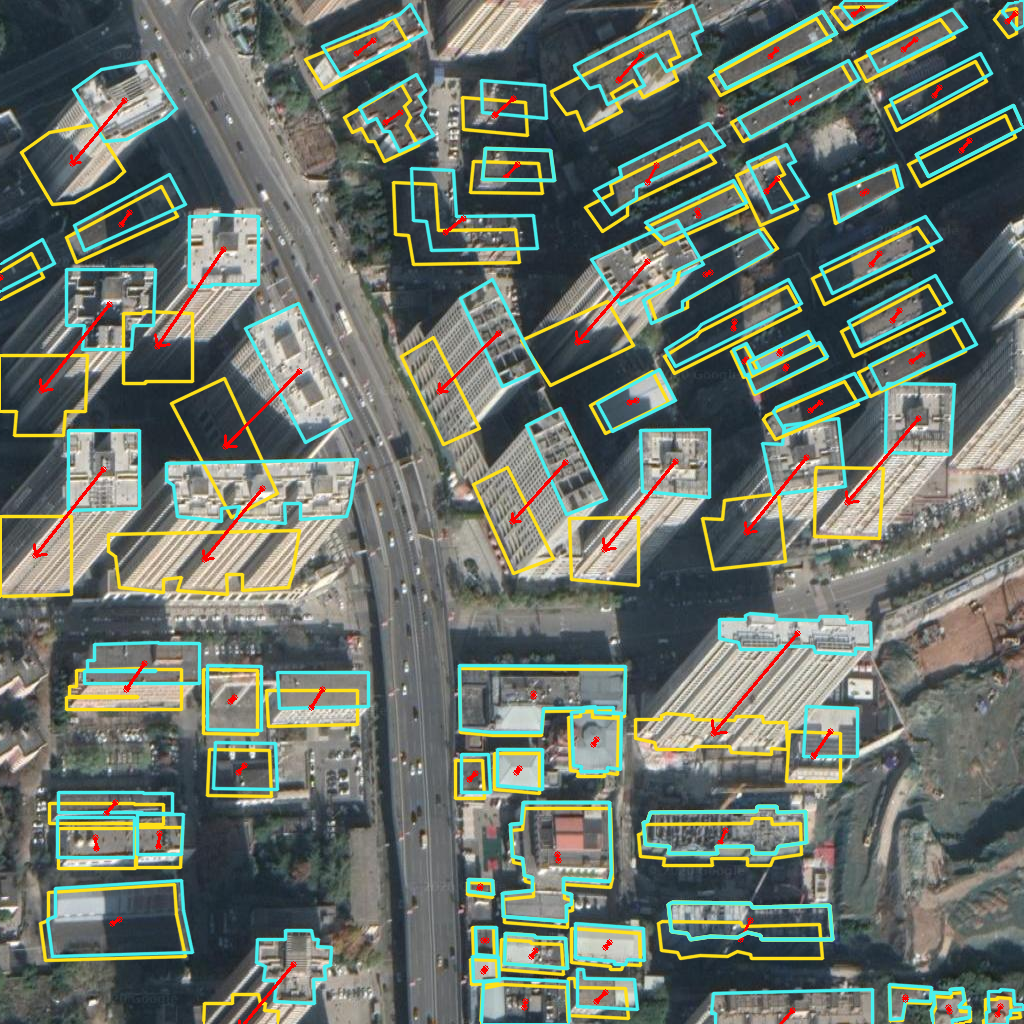}
	\end{tabular}
    \vspace{-4mm}
    \caption{Several annotated samples in BONAI dataset geo-located in different cities. For every building, its roof (blue) and footprint (yellow) are labeled with polygonal masks, and the offset vector (red) records the translation between them.}
    \label{fig:BONAI_vis}
    \vspace{-4mm}
\end{figure}

\subsection{BONAI Dataset}
\label{sec::dataset}

We build the BONAI dataset with a large quantity of off-nadir aerial images with spatial resolutions of $0.3$m and $0.6$m, mainly collected from Google Earth images\footnote{https://earth.google.com/} and Microsoft Virtual Earth images\footnote{http://www.microsoft.com/maps/} geo-located in six representative cities of China, \ie, {\em Shanghai, Beijing, Harbin, Jinan, Chengdu}, and {\em Xi'an}. It is worth noticing that although the images used in BONAI dataset are often RGB-rendered versions of original aerial images, the structure and appearance of the image content are always consistent and are feasible for recognition-oriented tasks~\cite{DOTA_2018_CVPR,DOTA_PAMI}.

As mentioned before, due to the observation that the contours of the roof and the footprint of a building in near-nadir images are often well overlapped, existing datasets for the BFE problem with near-nadir images, \eg,~\cite{WHU_2018_TGRS,SpaceNet_MVOI_2019_CVPR}, often label building footprints directly by their roofs.
In contrast, our BONAI dataset targets the BFE problem in off-nadir images and provides an instance-level annotation of every building. More precisely, for a building in BONAI dataset, we provide the masks of its roof and footprint in polygon formats as well as an associated offset vector from the roof toward the footprint. Fig.~\ref{fig:near_and_off_nadir} (c) illustrates an annotation example of a building in BONAI dataset.

\begin{table}[b!]
    \vspace{-3mm}
    \caption{The quantity of images and building instances contained by BONAI.}
    \label{tab:dataset_number}
    \vspace{-2.5mm}
	\centering
    \renewcommand{\arraystretch}{0.97}
    \renewcommand{\tabcolsep}{4.0mm}
    \begin{tabular}{crrr}
        \toprule
        BONAI dataset                               & City              & \#Image   & \#Instance  \\
        \midrule
        \multirow{4}{*}{Training Set}               & {\em Shanghai}        & 1,656     & 167,595  \\
                                                    & {\em Beijing}         & 684       & 36,932  \\
                                                    & {\em Chengdu}         & 72        & 4,448  \\
                                                    & {\em Harbin}          & 288       & 16,480  \\
        \midrule
        \multirow{2}{*}{Validation Set}             & {\em Shanghai}        & 228       & 16,747  \\
                                                    & {\em Jinan}           & 72        & 6,147  \\
        \midrule
        \multirow{2}{*}{Test Set}                & {\em Shanghai}        & 200       & 15,100  \\
                                                    & {\em  Xi'an}          & 100       & 5,489  \\
        \midrule
        Total                                       &      -                & 3,300     & 268,958 \\
        \bottomrule
    \end{tabular}
    \vspace{-3mm}
\end{table}

Noticing that the building footprints in off-nadir aerial images are usually heavily occluded, it is often impossible to directly annotate their accurate boundaries. Therefore, we first annotate its fully visible roof with a polygonal mask when labeling a building. We then create its footprint mask by translating the obtained roof mask to the footprint whose boundary is partially visible. The corresponding translation vector is finally recorded as the offset vector. Fig.~\ref{fig:BONAI_vis} shows some annotated images with building roof polygons in blue, footprint polygons in yellow, and offset vectors in red. 

Tab.~\ref{tab:dataset_number} presents the overview of the BONAI dataset, which contains $268,958$ buildings across $3,300$ images with size of $1024\times 1024$ pixels. The BONAI dataset is carefully split into three subsets, \ie, {\em Training Set}, {\em Validation Set}, and {\em Test Set}, such that the coverage areas of the images contained by one subset are geographically non-overlapped with those of others.

\subsection{Implementation Details}
Following the pipeline in Fig.~\ref{fig:SV-LOVE-framework}, we use ResNet-50~\cite{ResNet_2016_CVPR} pre-trained on the ImageNet with FPN~\cite{FPN_2017_CVPR} as the backbone. All the models are trained with a batch size of $32$ on $16$ NVIDIA Titan XP GPUs (with 12GB RAM)\footnote{Note that a single Titan XP GPU is sufficient to train and test our models. More GPUs can speed up the training and inference, while the influences on the model accuracy are negligible.}. We use $24$ epochs for training, with the learning rate starting from $0.02$ and decaying by a factor of $0.1$ at the $16^{th}$ and $22^{nd}$ epoch. The Stochastic Gradient Descent (SGD) with a weight decay of $0.0001$ and momentum of $0.9$ is used in all experiments. Mask R-CNN~\cite{Mask-R-CNN_2017_ICCV} is used as the basic architecture of the LOFT model unless specified otherwise. The batch sizes of the RPN and Fast R-CNN are set to be $512$ and $1024$, respectively, with a sampling ratio of 1/3 between the positives and negatives. The number of RPN proposals is set to $3000$, and we run the R-CNN head on these proposals, followed by Non-Maximum Suppression (NMS). The mask head and offset head are then applied to $512$ building boxes with the highest scores. All models are built in PyTorch.

\begin{figure*}[!t]
    \renewcommand{\tabcolsep}{0.4mm}
    \renewcommand{\arraystretch}{0.50}
    \newcommand{\firstimage}{L18_104544_210392__0_1024}
    \newcommand{\secondimage}{L18_104512_210416__0_1024}
    \newcommand{\thirdimage}{L18_104536_210432__1024_0}
    \newcommand{\fourthimage}{shanghai_arg__L18_107192_219496__1024_1024}
    \centering
	\begin{tabular}{ccccc}
	
	\includegraphics[width=0.195\linewidth]{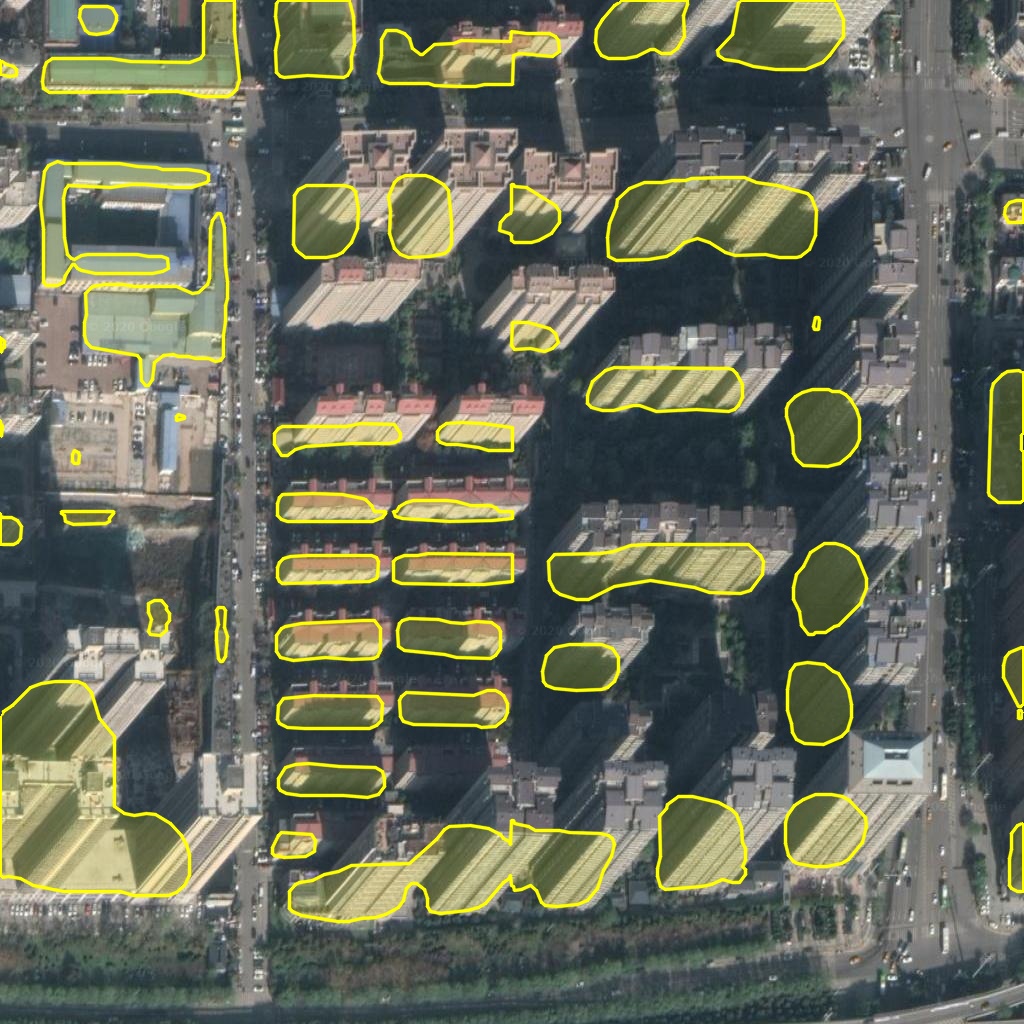} &
    \includegraphics[width=0.195\linewidth]{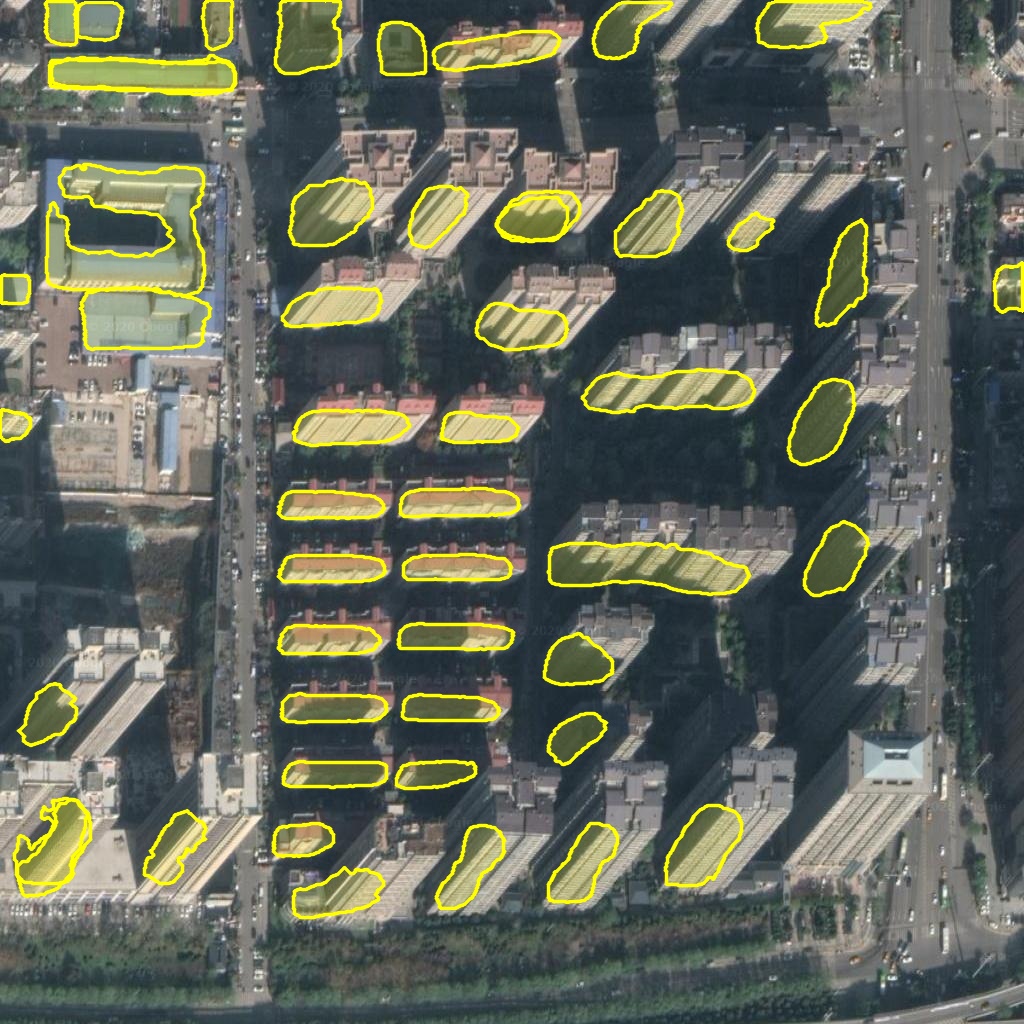} &
    \includegraphics[width=0.195\linewidth]{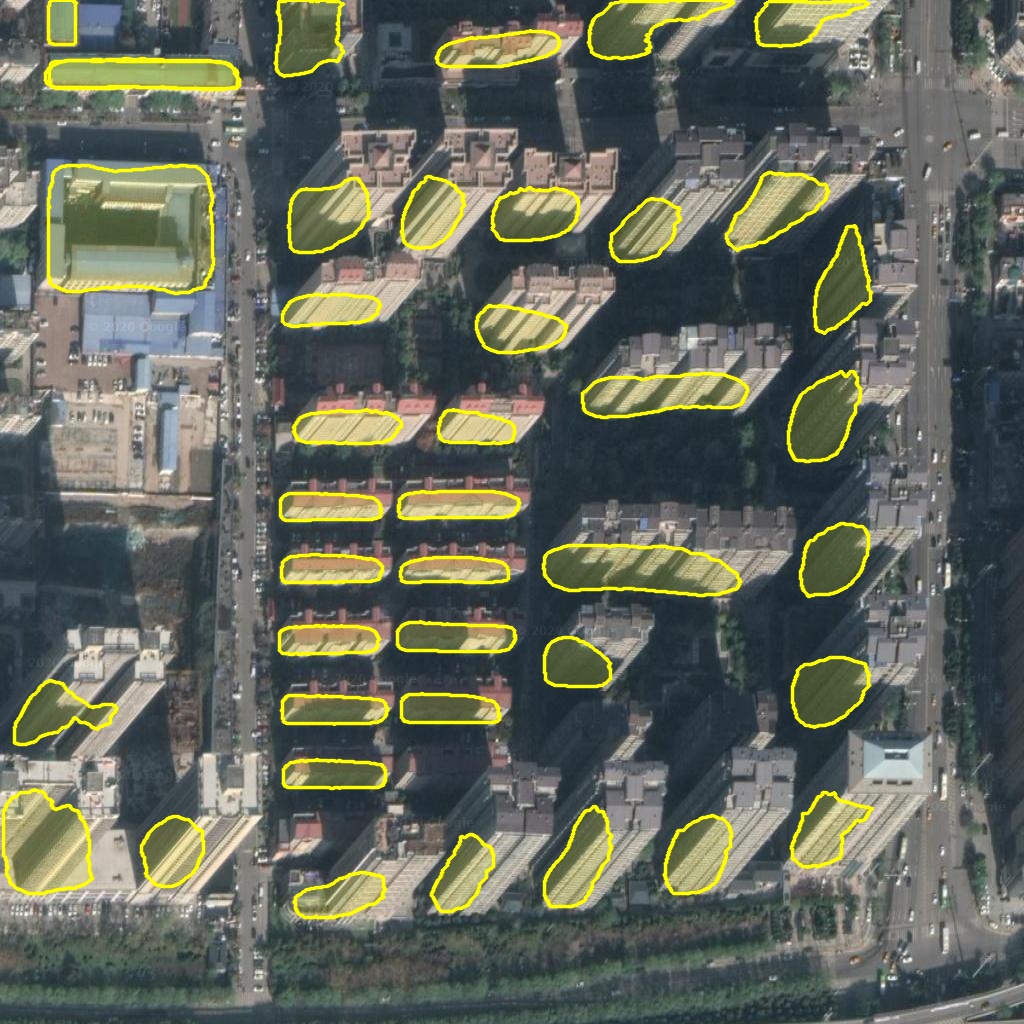} &
    \includegraphics[width=0.195\linewidth]{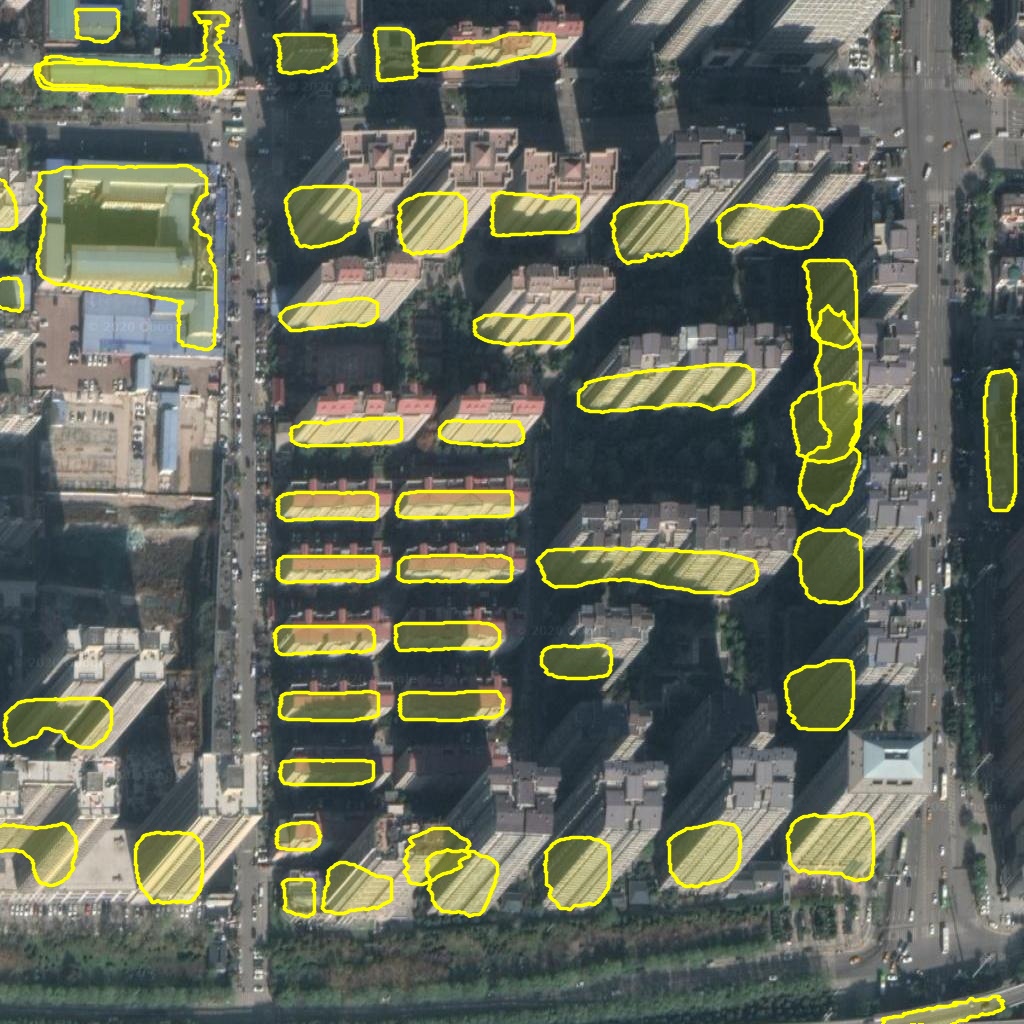} &
    \includegraphics[width=0.195\linewidth]{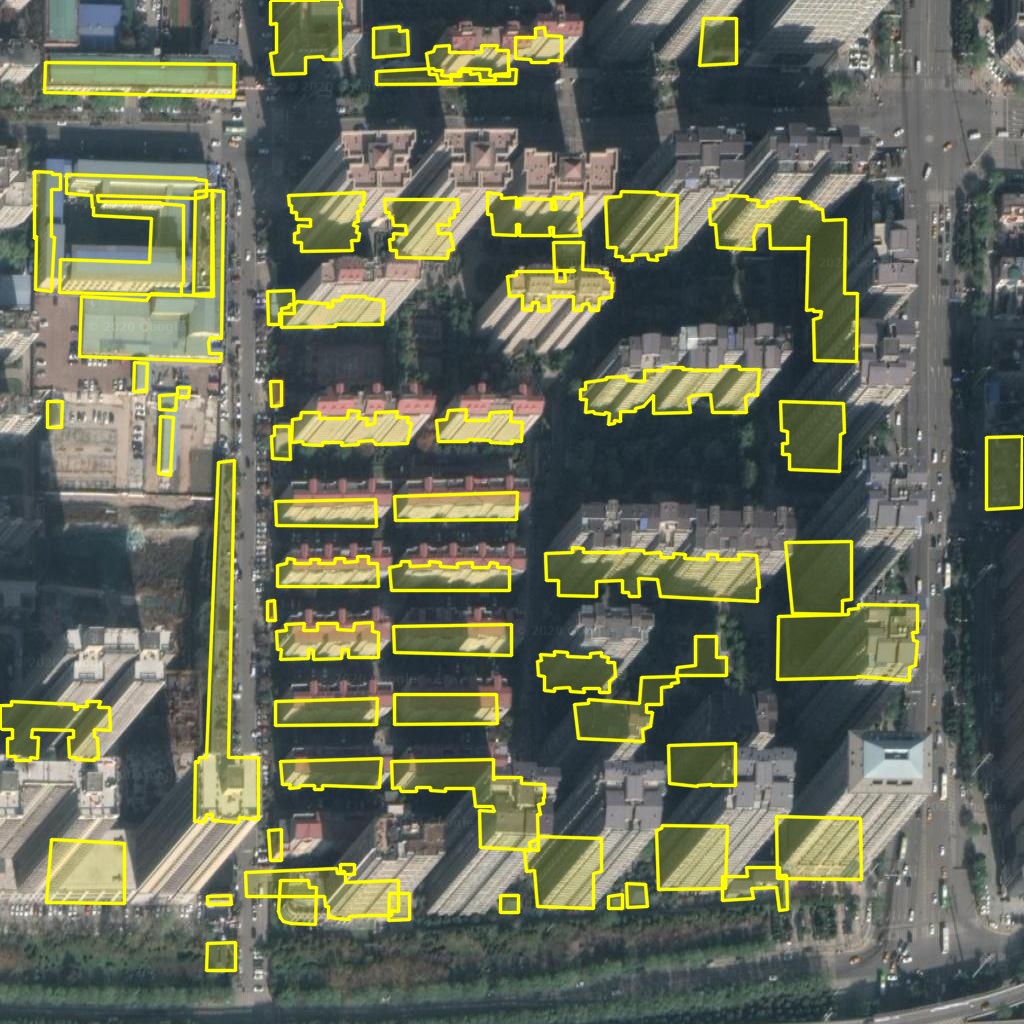}
	\\
	\includegraphics[width=0.195\linewidth]{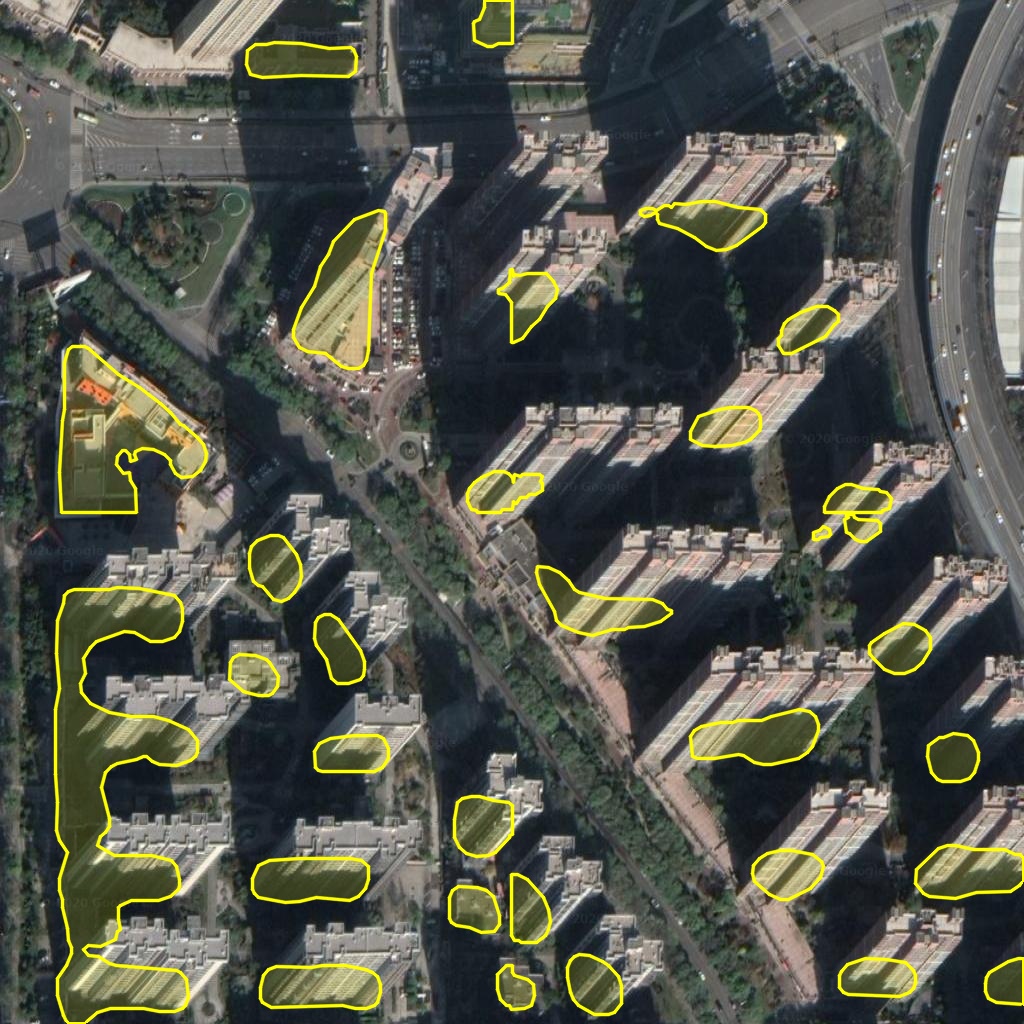} &
    \includegraphics[width=0.195\linewidth]{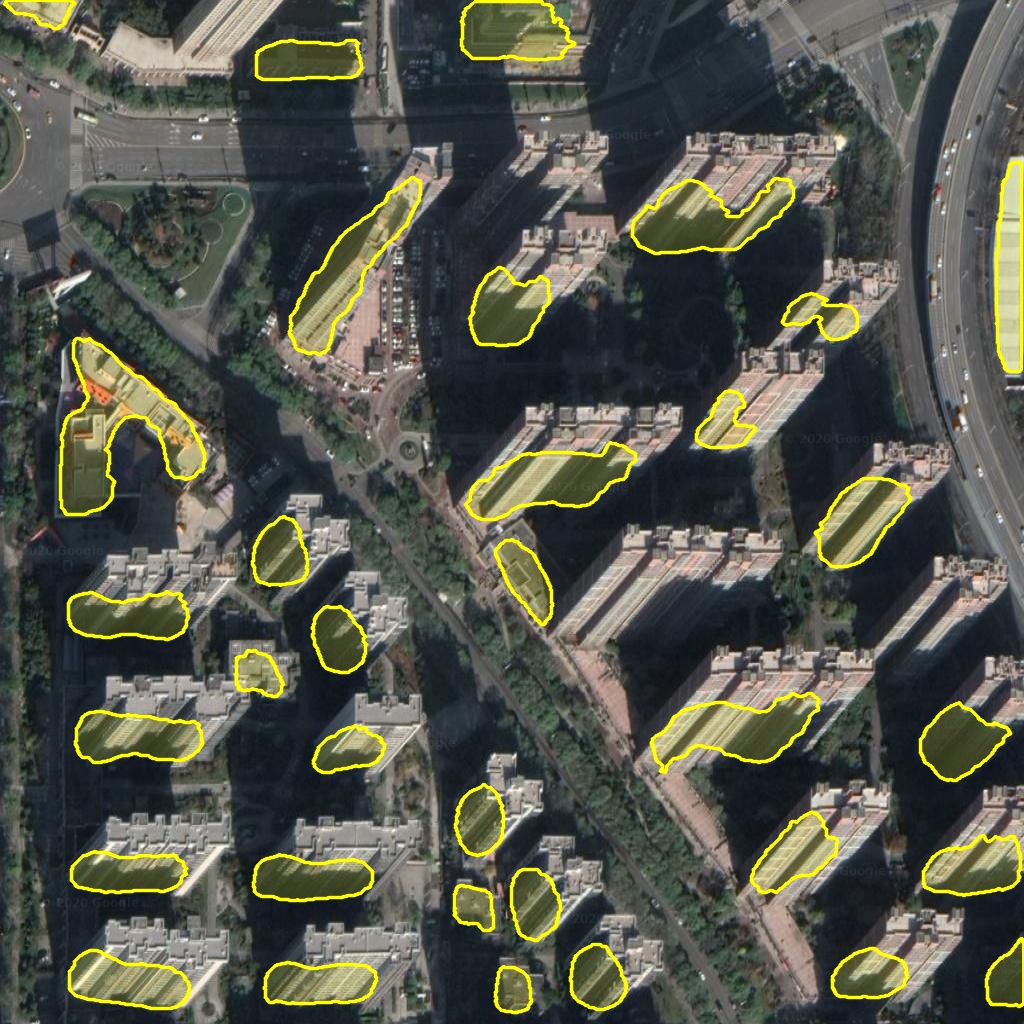} &
    \includegraphics[width=0.195\linewidth]{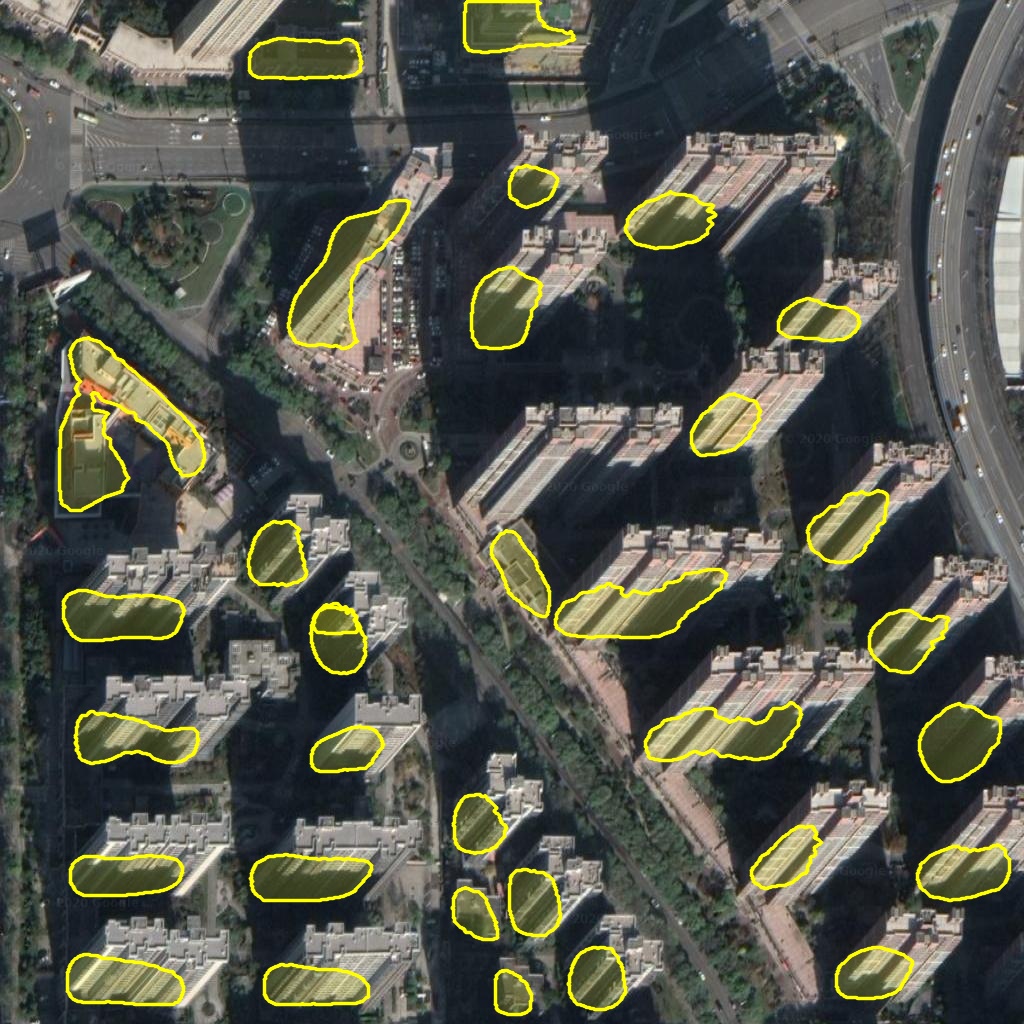} &
    \includegraphics[width=0.195\linewidth]{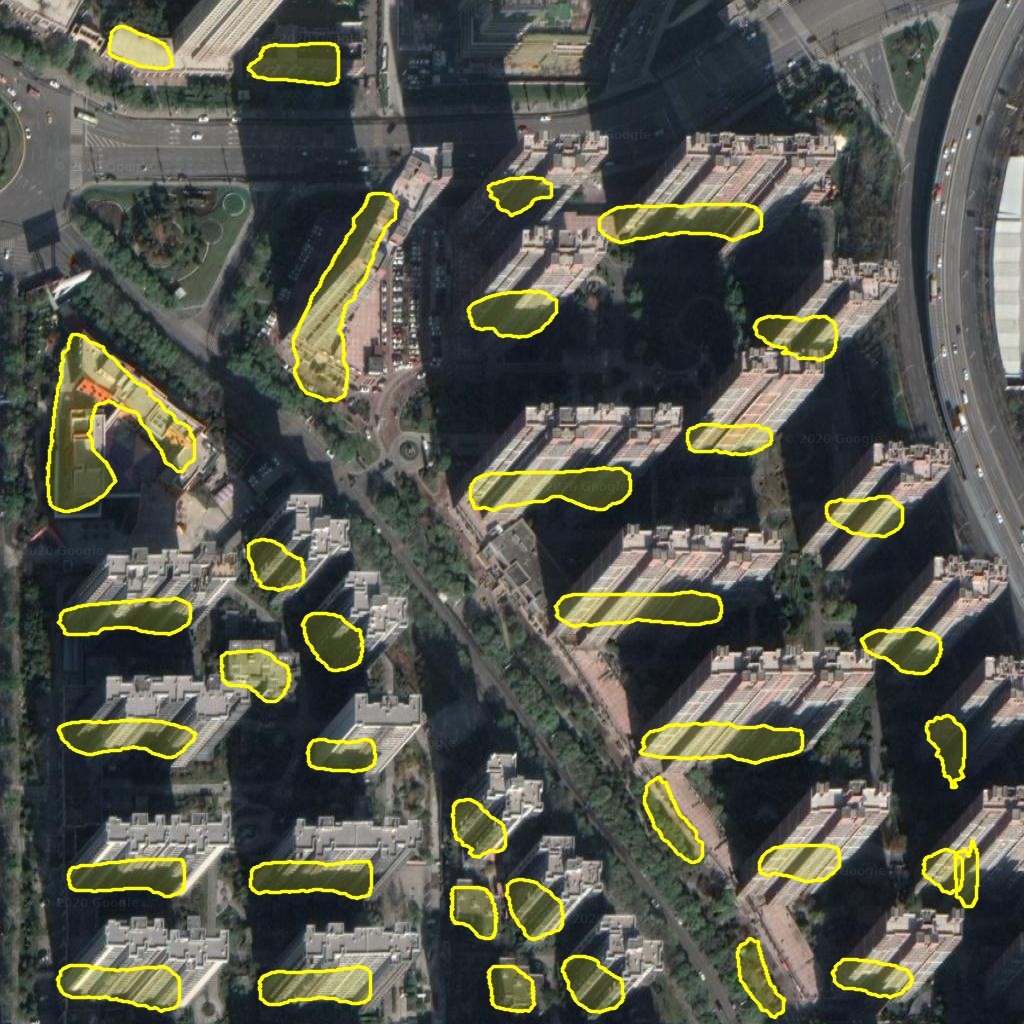} &
    \includegraphics[width=0.195\linewidth]{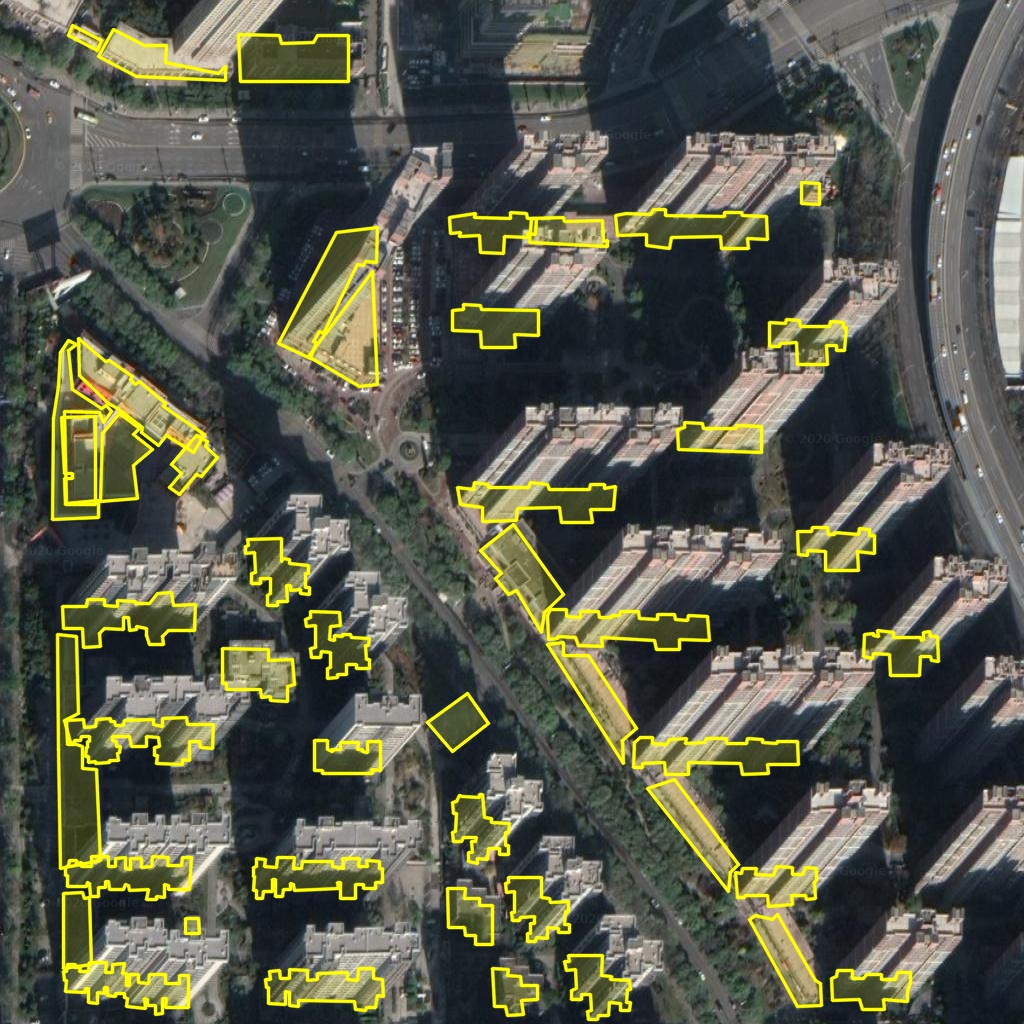}
	\\
	\includegraphics[width=0.195\linewidth]{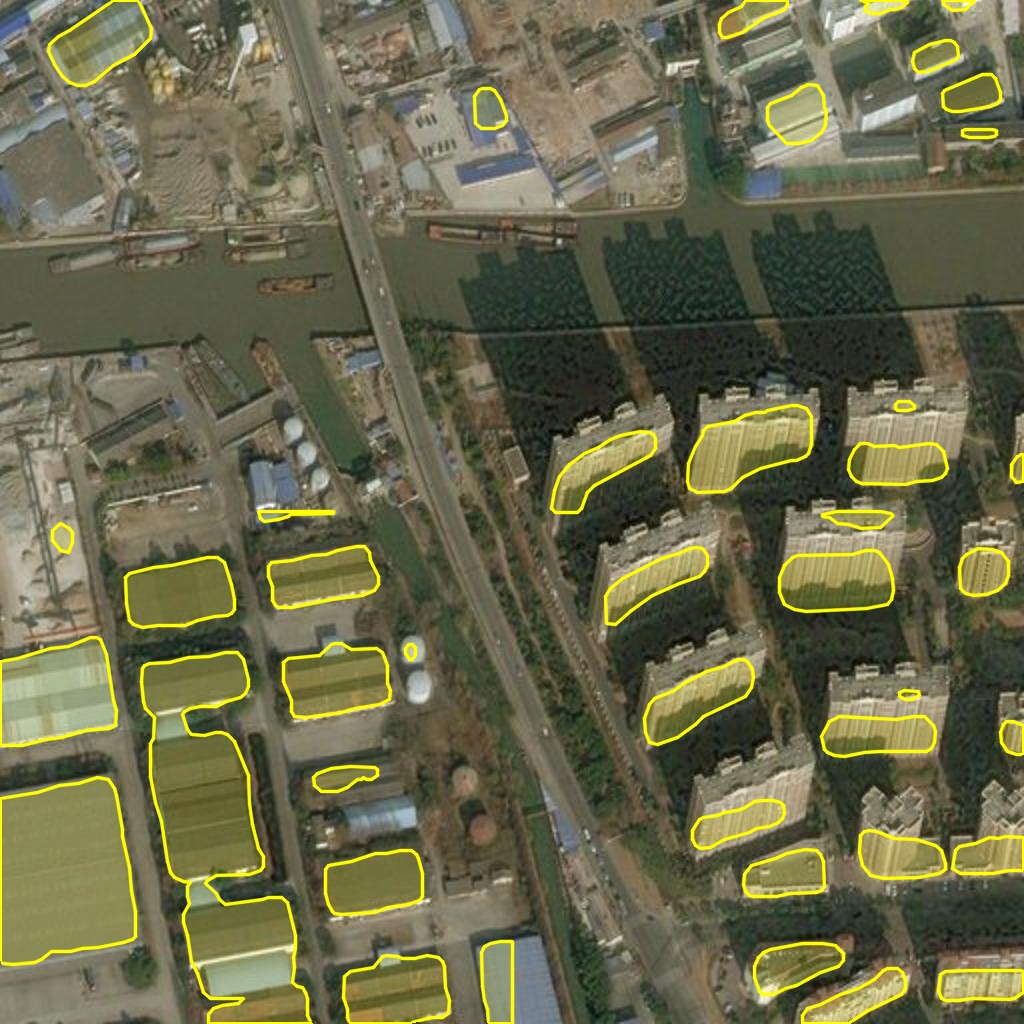} &
    \includegraphics[width=0.195\linewidth]{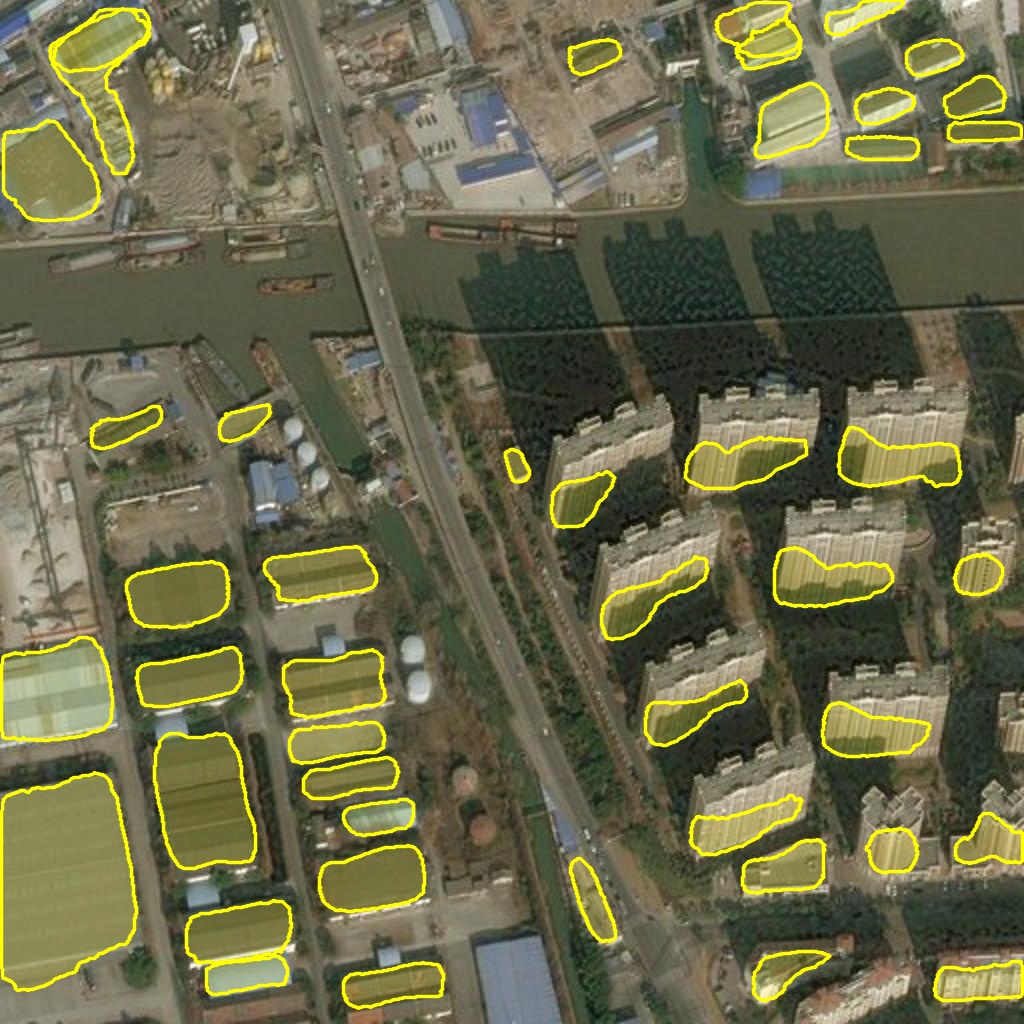} &
    \includegraphics[width=0.195\linewidth]{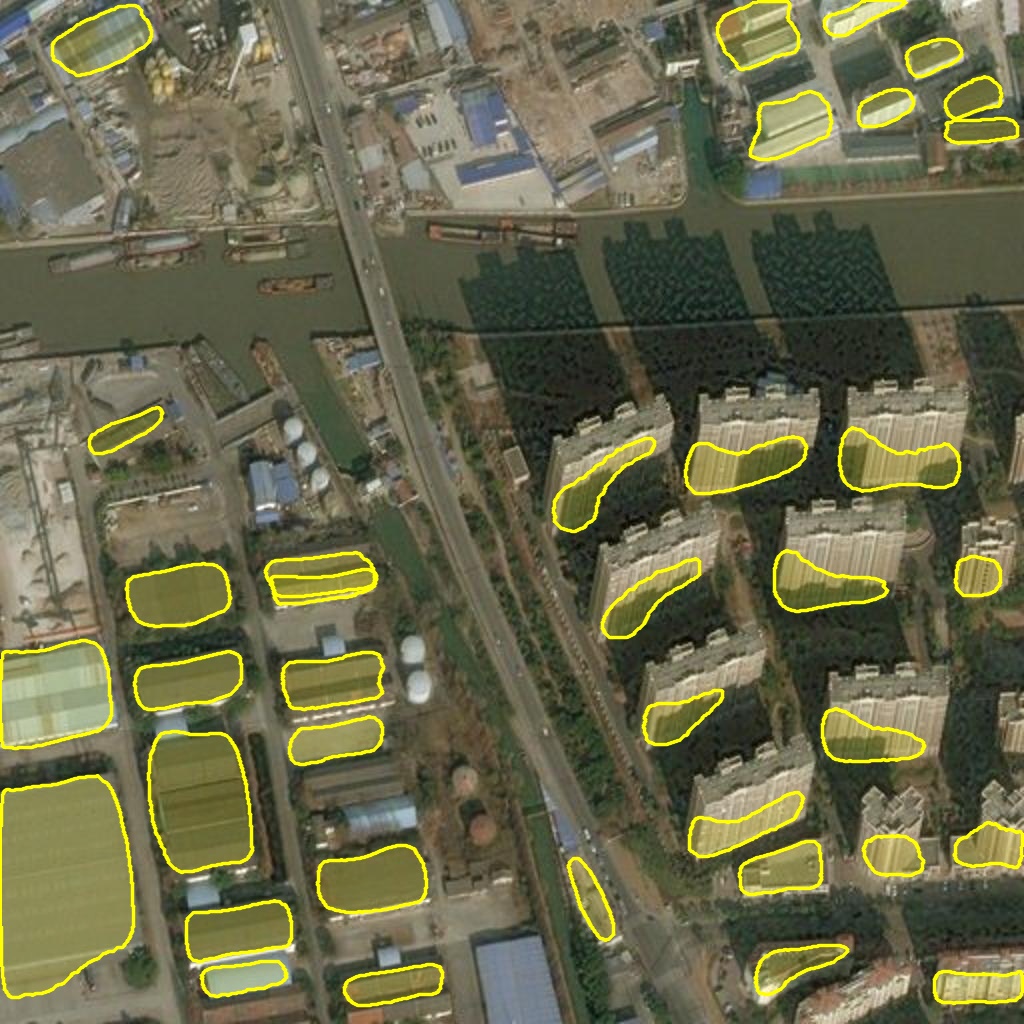} &
    \includegraphics[width=0.195\linewidth]{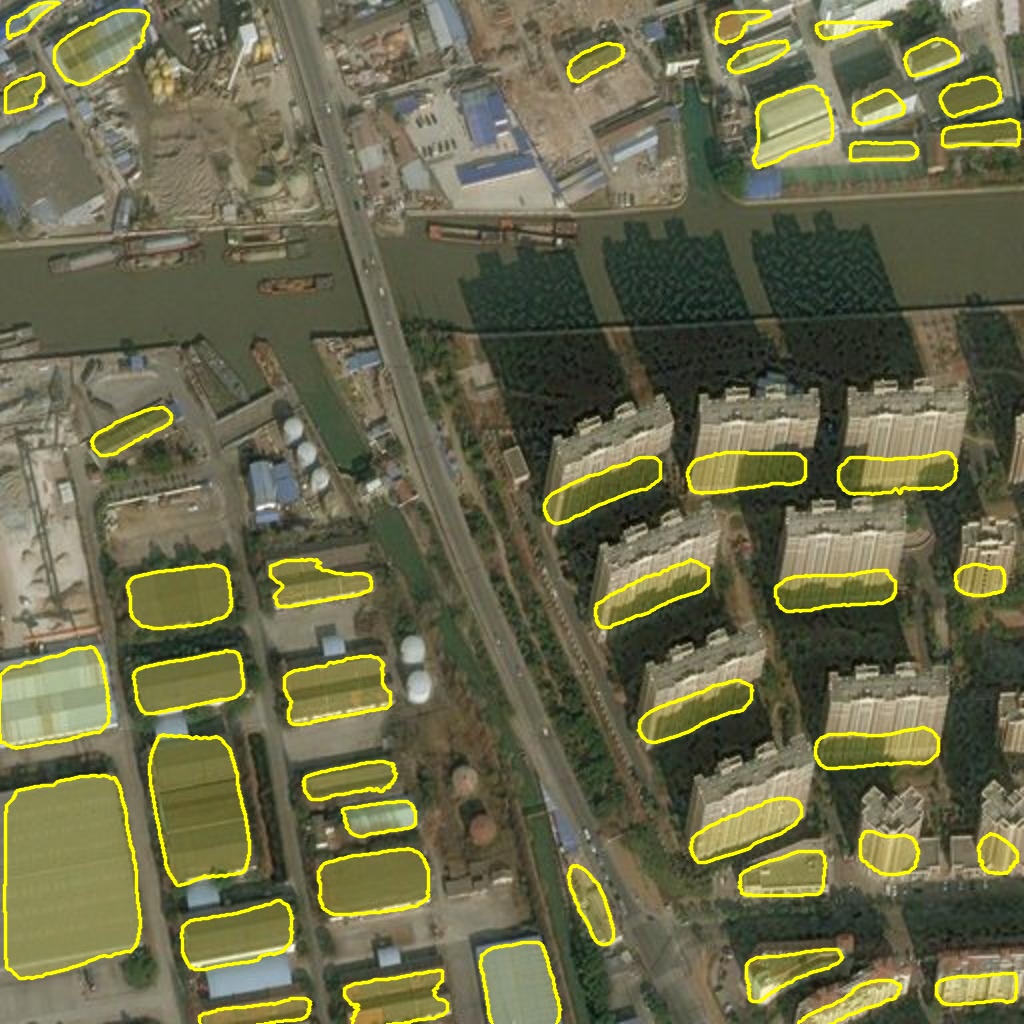} &
    \includegraphics[width=0.195\linewidth]{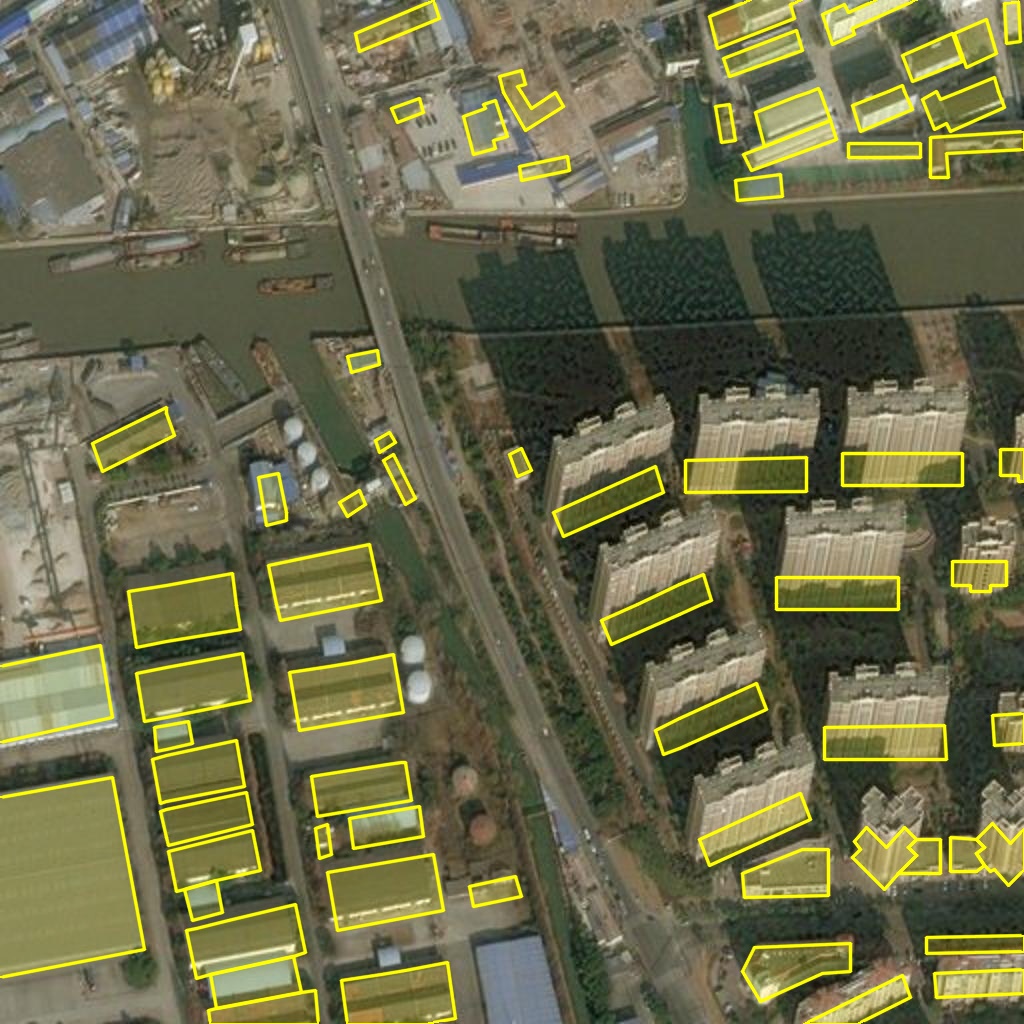}
    \\
	\footnotesize{MAP-Net~\cite{MAP-Net_2020_TGRS}} & \footnotesize{Mask R-CNN~\cite{Mask-R-CNN_2017_ICCV}} & \footnotesize{CM R-CNN~\cite{HTC_2019_CVPR}} & \footnotesize{Ours} & \footnotesize{}Ground Truth
	\end{tabular}
	
	\caption{Qualitative comparison between the prior arts of BFE and our method on BONAI dataset. Best viewed in color and zoomed-in view.}
	\vspace{-3mm}
	\label{fig:offset-rcnn-vis}
\end{figure*}

\subsection{Evaluation Protocols}

A core step of evaluating a BFE method is to instance-wisely match the predicted footprint to the ground truth with a specified metric. To assess the localization quality and boundary quality simultaneously, two {\em Intersection over Union} (IoU)-based segmentation evaluation measures, \ie, Mask IoU and Boundary IoU~\cite{Boundary_IoU_2021_CVPR}, are adopted in our experiments.

The most commonly used evaluation metrics for BFE task are {\em Precision, Recall}, and {\em F1-Score} with a Mask IoU threshold of $0.5$. These metrics mainly measure the localization quality since Mask IoU is insensitive to the boundaries~\cite{Boundary_IoU_2021_CVPR}. However, the boundary quality is another crucial factor in evaluating of building footprint extractors. Therefore, we employ a recently proposed metric called Boundary AP$_{50}$ (AP$_{50}^B$)~\cite{Boundary_IoU_2021_CVPR} to obtain a more reliable evaluation of boundary performance. AP$_{50}^B$ is proposed by replacing Mask IoU by the Boundary IoU~\cite{Boundary_IoU_2021_CVPR} in Average Precision (AP) metric with a Boundary IoU threshold of $0.5$. The Boundary IoU can better reveal the improvements in boundary quality that Mask IoU  generally ignores. Note that since the BFE task is also an instance segmentation task, the AP metric can be used to evaluate the performance of BFE methods. Finally, the F1-Score is exploited as the main metric to be consistent with other works in literature.

\subsection{Main Results}
\label{sec:main_results}

\begin{table}[t!]
    \caption{Quantitative comparison of the baselines and our methods on the test set of BONAI dataset. Bold and underline fonts indicate the best and the second best performances for each metric (\%).}
    \vspace{-2mm}
    \label{tab:main_results}
    \renewcommand{\arraystretch}{1.1}
    \renewcommand{\tabcolsep}{2mm}
    \centering
    \begin{tabular}{lcccc}
        \toprule
        Method                                      & F1-Score & Precision     & Recall & AP$_{50}^B$\\
        \midrule
        MAP-Net~\cite{MAP-Net_2020_TGRS}            & 56.92        & 58.01        & 55.93 & 30.30\\
        PANet~\cite{PANet_2018_CVPR}                & 58.06        & 59.26        & 56.91 & 45.30\\
        Mask R-CNN~\cite{Mask-R-CNN_2017_ICCV}      & 58.12        & 59.26        & 57.03 & 45.80\\
        HRNetv2-W32~\cite{HRNet_2019_arXiv} & 60.81 & 61.20 & \underline{60.42} & \underline{50.10}\\
        CM R-CNN~\cite{HTC_2019_CVPR}     & \underline{60.94}        & \textbf{67.09}        & 55.83 & 46.50 \\
        Ours     & \textbf{64.31}  & \underline{63.37} & \textbf{65.29} & \textbf{53.40}\\
        \bottomrule
    \end{tabular}
    \vspace{-2mm}
\end{table}

We evaluate the BFE performance of our proposed method on two datasets, \ie, the BONAI dataset containing off-nadir images and the commonly-used WHU Building dataset~\cite{WHU_2018_TGRS} that 
only has near-nadir images.

\textbf{Results on Off-nadir Images.} We compare the performance of our LOFT with the state-of-the-art instance segmentation methods which are used for the {\em SpaceNet Building Detection Challenge}\footnote{https://spacenetchallenge.github.io/} on the BONAI dataset. The results are reported in Tab.~\ref{tab:main_results}. One can see that our proposed method (LOFT w/ FOA) achieves an absolute improvement of $3.37$ points in terms of F1-score when compared with the state-of-the-art instance segmentation method Cascade Mask R-CNN (CM R-CNN)~\cite{HTC_2019_CVPR}. Besides, our method performs much better than MAP-Net~\cite{MAP-Net_2020_TGRS}, indicating the effectiveness and superiority of our method on off-nadir images. We visualize some representative footprint extraction results in Fig.~\ref{fig:offset-rcnn-vis}, which qualitatively demonstrates the superiority of LOFT to its counterparts in terms of the accuracy of position and shape. Methods that directly extract footprints tend to mistake the building facades as the building footprints, since the network needs to learn the structural information of the occluded boundaries implicitly when training. Instead, for the LOFT scheme, the shape predictions of the building footprints are usually correct for most buildings as it extracts the building footprints indirectly by predicting the fully visible building roofs and the corresponding offsets.

\textbf{Results on Near-nadir Images.}  To further evaluate the generalization ability of our method, we train the LOFT with or without offset head on the WHU Building dataset (WHU dataset)~\cite{WHU_2018_TGRS} which is designed to evaluate the BFE methods for near-nadir images. In the experiments, we train and evaluate our method on the aerial subsets, which consist of more than $187,000$ building instances across $8,188$ aerial image tiles with $512\times 512$ pixels. We conduct our experiments with the same setting as in~\cite{WHU_2018_TGRS}. Note that we set offset vector as $[0, 0]$ for each building in the experiment of the LOFT with offset head since the WHU dataset only contains near-nadir images. As a result, the F1-Scores of the LOFT with and without offset head both achieve $91.97\%$ which is comparable to the $91.76\%$ reported in~\cite{WHU_2018_TGRS} with the same Mask R-CNN architecture, which implies our LOFT with offset head can also work well for near-nadir images. Therefore our proposed LOFT is a unified BFE method for both off-nadir and near-nadir images, as the offset vectors tend to be zero in near-nadir images.

\subsection{Ablation Study}
\label{sec:ablation_study}

We also conduct a series of experiments to investigate the function of each component in the proposed method. The detailed comparisons are given in the following.

\begin{table}[t]
    \caption{The influence of offset head on BONAI dataset (\%).}
    \label{tab:influence_offset}
    \vspace{-2mm}
    \renewcommand{\tabcolsep}{2.0mm}
    \renewcommand{\arraystretch}{1.1}
    \centering
    \begin{tabular}{lcccc}
        \toprule
        Method                        & F1-Score & Precision & Recall & AP$_{50}^B$\\
        \midrule
        Mask R-CNN~\cite{Mask-R-CNN_2017_ICCV}      			  & 58.12     & 59.26           & 57.03 & 45.80\\
        
        Mask R-CNN + Offset       	  & \textbf{61.78}    & \textbf{60.87}         & \textbf{62.72} & \textbf{50.10}\\
        \midrule
        PANet~\cite{PANet_2018_CVPR}                         & 58.06 & 59.26 & 56.91 & 45.30 \\
        PANet + Offset                & \textbf{62.15} & \textbf{61.33} & \textbf{62.99} & \textbf{49.90}\\
        \midrule
        HRNetv2-W32~\cite{HRNet_2019_arXiv}            & 60.81 & 61.20 & 60.42 & 50.10\\
        HRNetv2-W32 + Offset   & \textbf{63.16} & \textbf{62.31} & \textbf{64.03} & \textbf{52.00}\\
        \midrule
        CM R-CNN~\cite{HTC_2019_CVPR}            & 60.94 & 67.09 & 55.83 & 46.50\\
        CM R-CNN + Offset   & \textbf{63.73} & \textbf{68.29} & \textbf{59.74} & \textbf{48.70} \\
        \bottomrule
    \end{tabular}
    \vspace{-3mm}
\end{table}

\textbf{Influence of the Offset Head.} We apply the offset head to typical top-down instance segmentation methods, \ie Mask R-CNN~\cite{Mask-R-CNN_2017_ICCV}, PANet~\cite{PANet_2018_CVPR}, HRNetv2-W32~\cite{HRNet_2019_arXiv}, and CM R-CNN~\cite{HTC_2019_CVPR}, to verify the effectiveness of the offset head. For the sake of fairness, all hyperparameters are strictly consistent. Tab.~\ref{tab:influence_offset} shows the comparison results. It can be observed that the offset head can improve the performance of Mask R-CNN, PANet, HRNetv2-W32, and CM R-CNN by $3.66$, $4.09$, $2.35$, and $2.79$ points in terms of F1-Score, respectively, implying the effectiveness of the offset-based method for BFE in off-nadir aerial images. It is worth noting that, in addition to the methods in Tab.~\ref{tab:influence_offset}, the offset head can also be applied to other instance segmentation methods.

\begin{figure}[!t]
    \renewcommand{\tabcolsep}{0.4mm}
    \renewcommand{\arraystretch}{0.7}
    \newcommand{\firstimage}{L18_104512_210416__0_1024}
    \newcommand{\secondimage}{shanghai_arg__L18_107160_219496__0_1024}
    \centering
	\begin{tabular}{ccc}
	
	\includegraphics[width=0.32\linewidth]{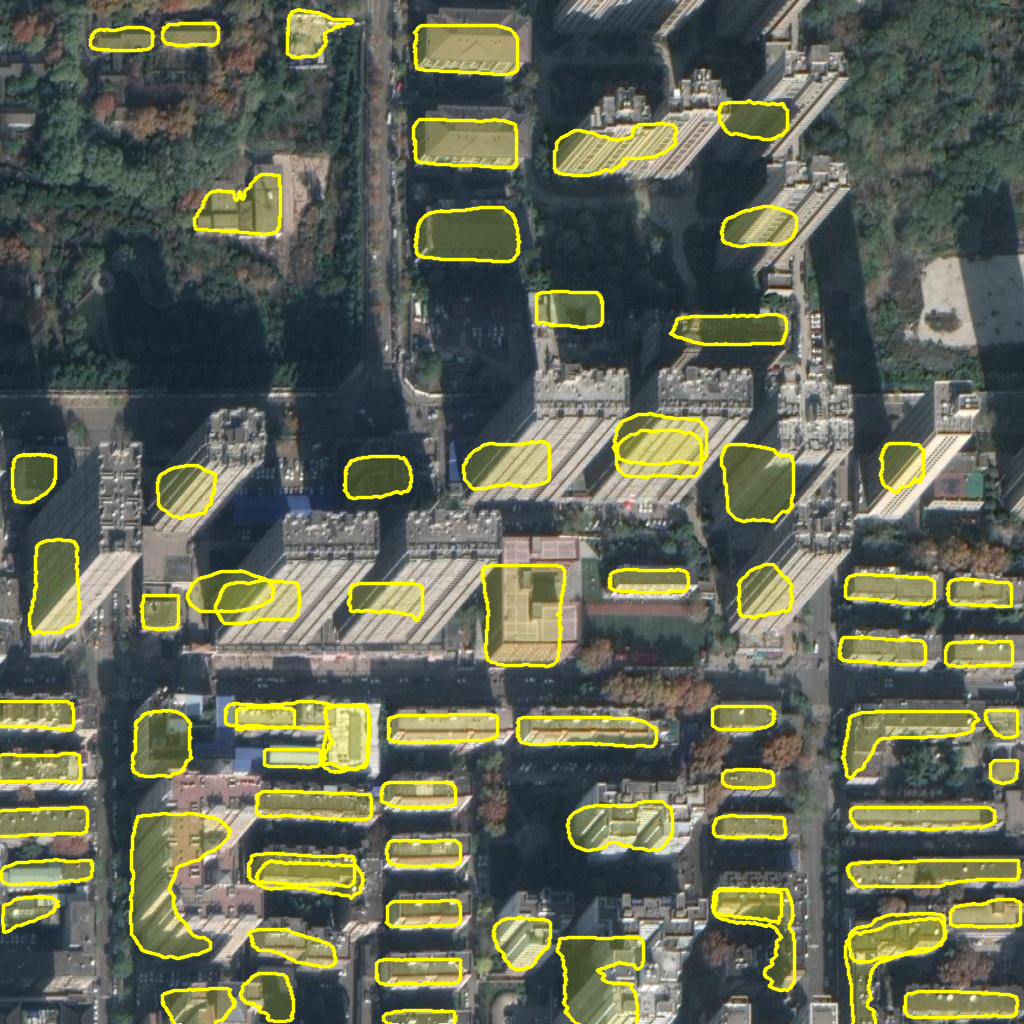} &
    \includegraphics[width=0.32\linewidth]{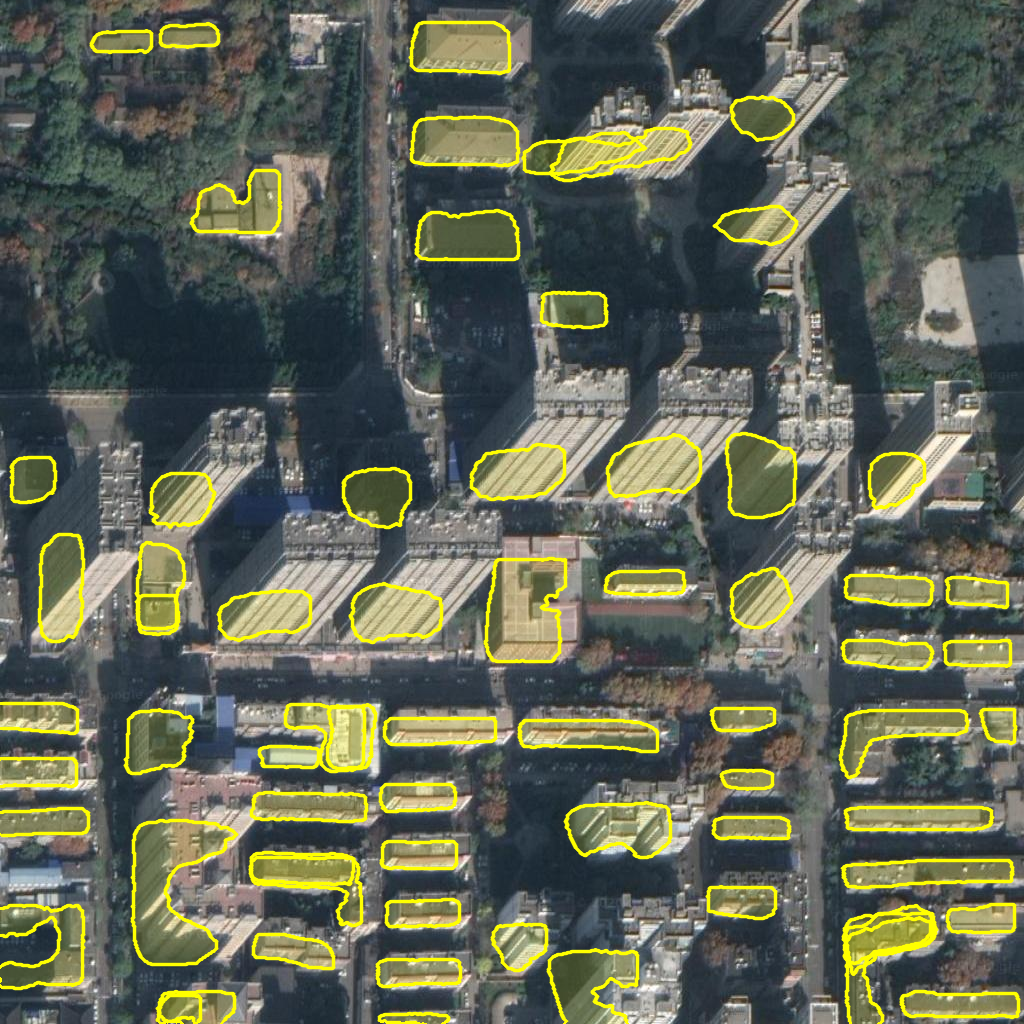} &
    \includegraphics[width=0.32\linewidth]{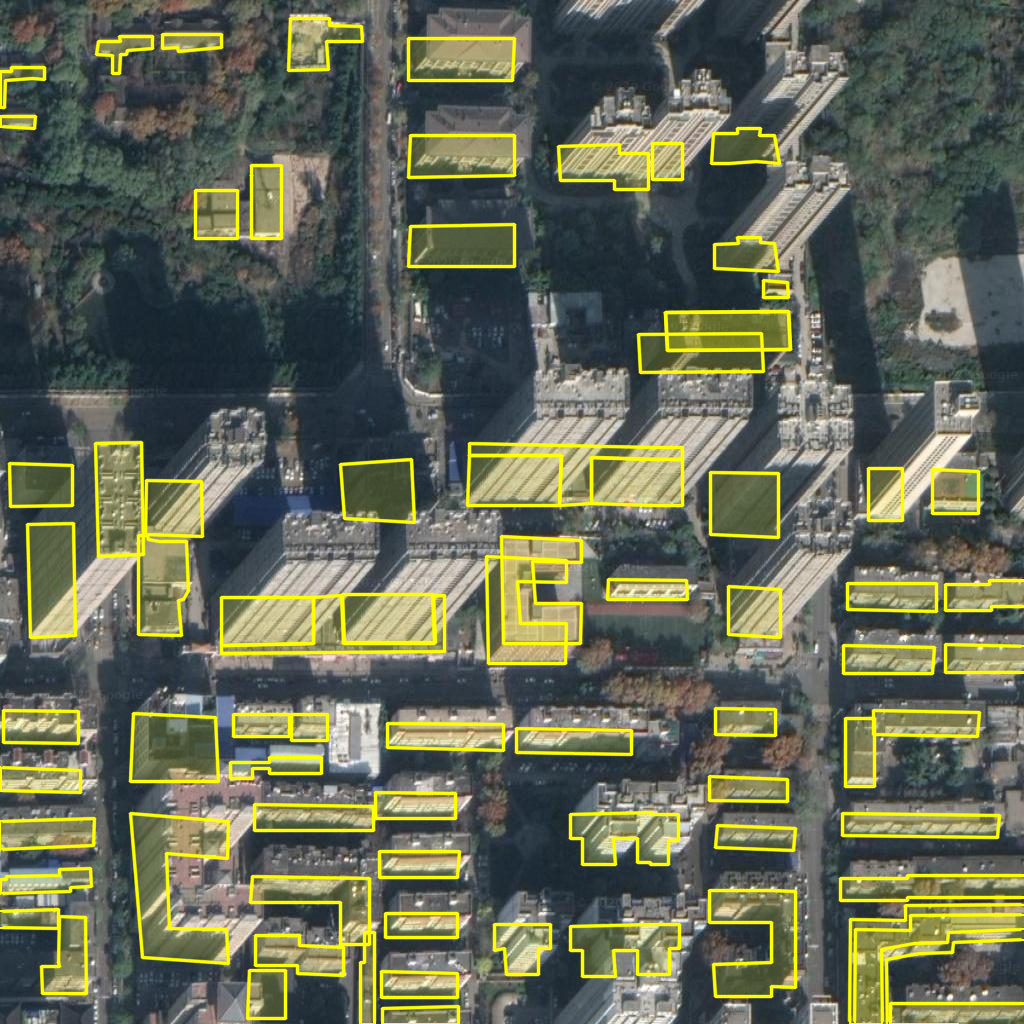}
	\\
	\includegraphics[width=0.32\linewidth]{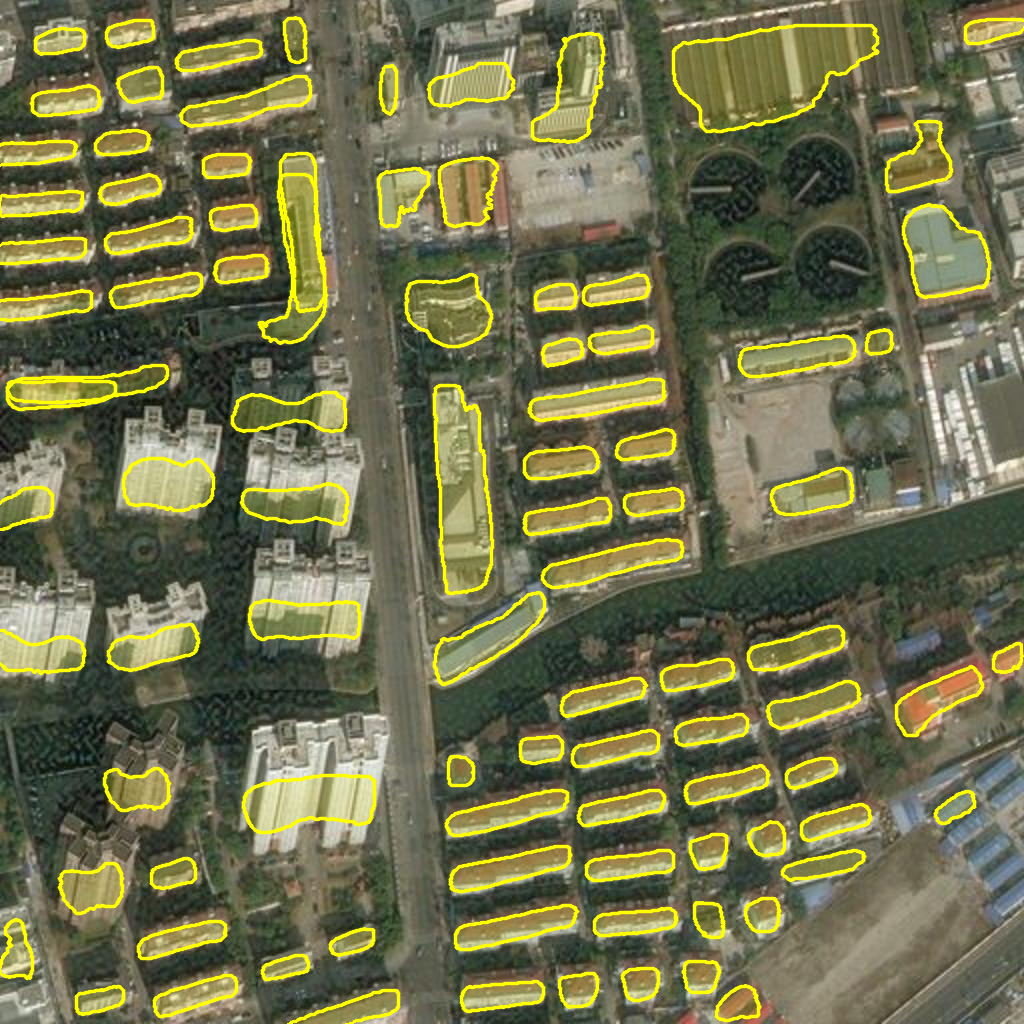} &
    \includegraphics[width=0.32\linewidth]{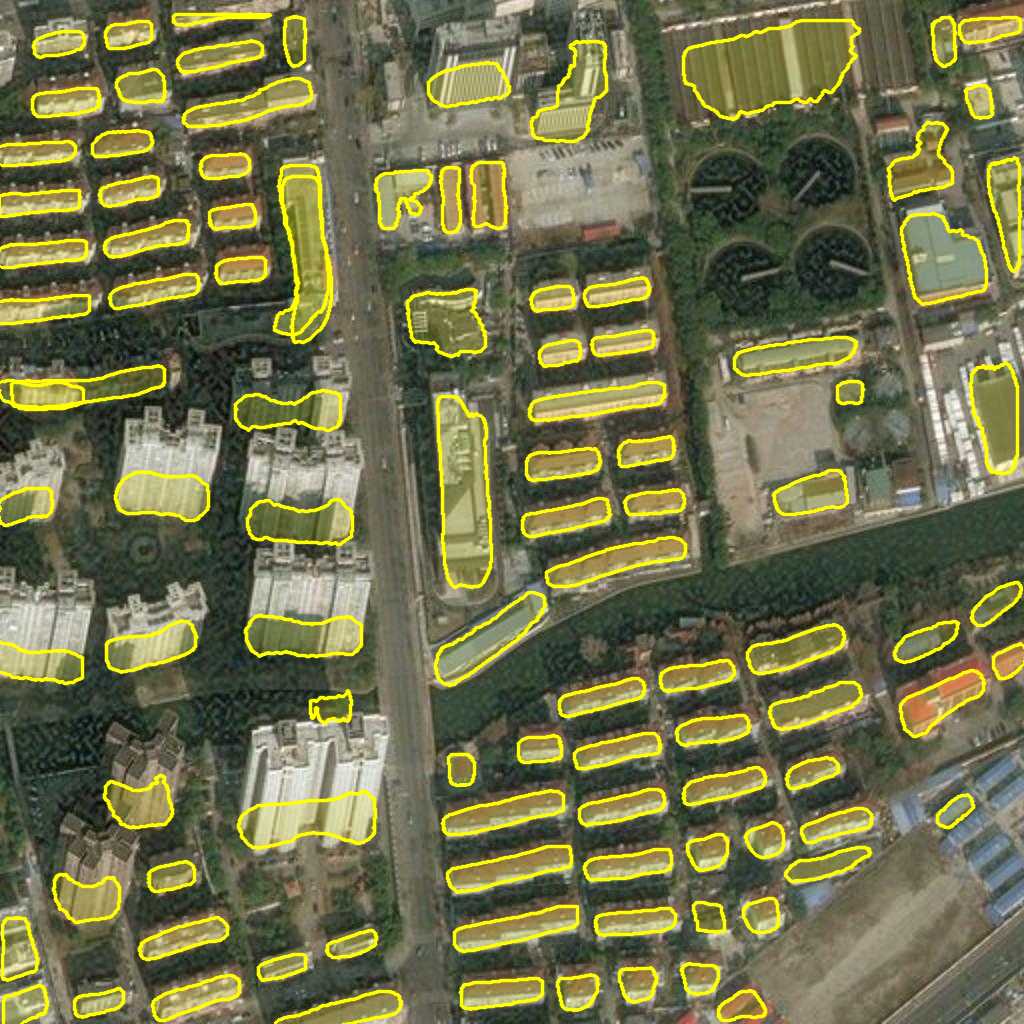} &
    \includegraphics[width=0.32\linewidth]{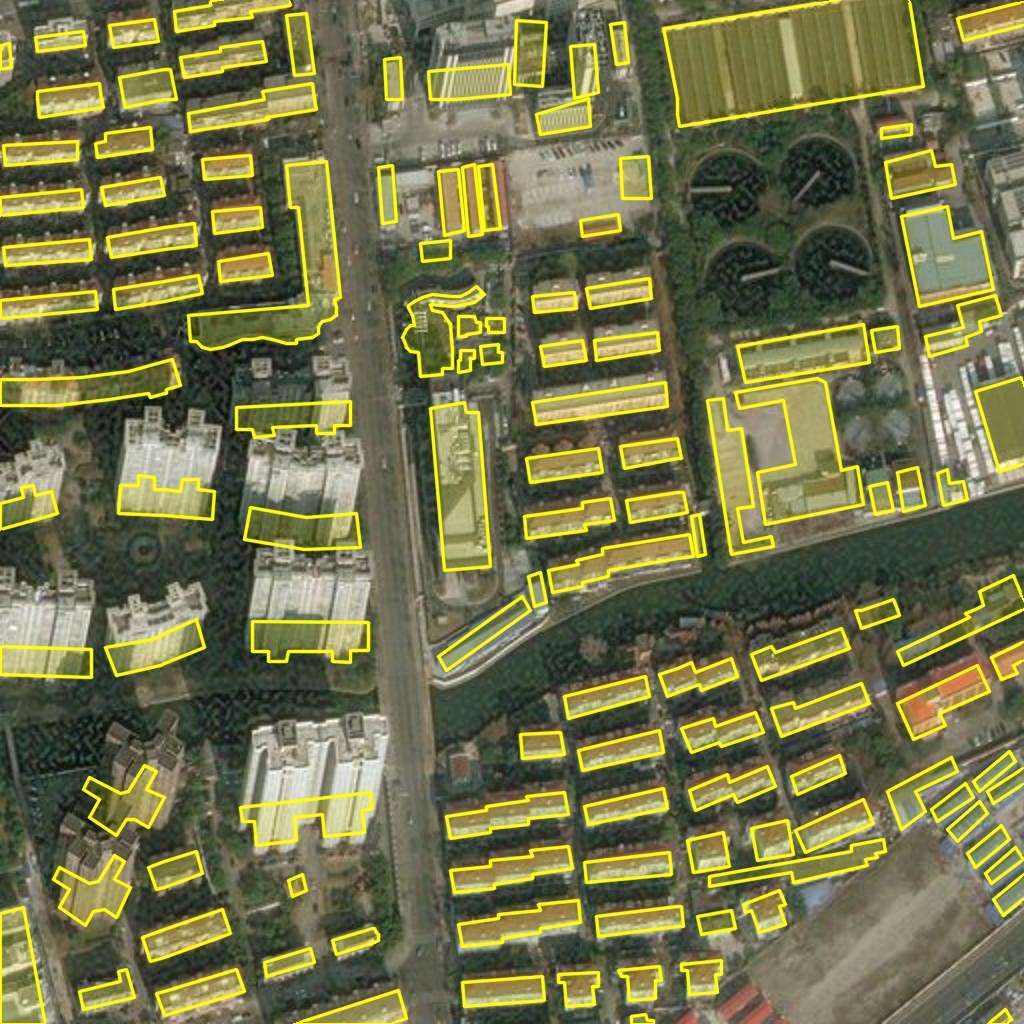}
    \\
	\footnotesize{LOFT} & \footnotesize{LOFT w/ FOA} & \footnotesize{Ground Truth}
	\end{tabular}
	
	\caption{Qualitative comparison between LOFT and LOFT w/ FOA. Best viewed in color and zoomed-in view.}
	\vspace{-3mm}
	\label{fig:foa-vis}
\end{figure}

\textbf{Influence of the FOA Module.} We also compare the performance of the LOFT with and without the use of the FOA module. The experimental results are illustrated in Tab.~\ref{tab:influence_foa}. It shows that the FOA module brings noticeable gain in contrast to LOFT without the FOA module, which improves the footprint F1-Score from $61.78\%$ to $64.31\%$. Besides, we can find that the roof F1-Scores of LOFT and LOFT w/ FOA are almost the same, indicating that the FOA module only improves the accuracy of offset prediction. Therefore, in the case of high offset prediction accuracy, we can easily utilize the better instance segmentation methods to predict the building roof for further improving the accuracy of building footprint extraction. In Fig.~\ref{fig:foa-vis}, we present the qualitative results on the BONAI dataset, where the proposed FOA generates more accurate offset vectors.

\begin{table}
\caption{The influence of the FOA module on BONAI dataset (\%).}
\label{tab:influence_foa}
\vspace{-2mm}
\renewcommand{\tabcolsep}{1.0mm}
\renewcommand{\arraystretch}{1.1}
\centering
\begin{tabular}{cc|cccc}
    \toprule
    Item & Method & F1-Score & Precision & Recall & AP$_{50}^B$\\
    \midrule
    \multirow{2}{*}{Roof} & LOFT & 67.17 & 65.49 & 68.95 & 59.60 \\
                         & LOFT w/ FOA & \textbf{67.25} & \textbf{65.55} & \textbf{69.03} & \textbf{60.80}\\
    \midrule
    \multirow{2}{*}{Footprint} & LOFT & 61.78 & 60.87 & 62.72 & 50.10\\
                       & LOFT w/ FOA & \textbf{64.31} & \textbf{63.37} & \textbf{65.29} & \textbf{53.40}\\
    \bottomrule
\end{tabular}
\vspace{-3mm}
\end{table}

\textbf{Influence of the Rotation Angles in FOA.} As mentioned in Sec.~\ref{sec::methodology}, the FOA module will transform instance-level building features and then predict multiple offsets according to the rotation angle set. Hence, we study the influence of different rotation angle sets including typical angle combinations, \ie, $\{0\}, \{0, \pi/2 \}, \{0, \pi/2, \pi\}, \{0, \pi/2, \pi, 3\pi/2\}$. Tab.~\ref{tab:offset_rcnn_foa} reports the results with different rotation angle sets,  implying that the footprint extraction performance can be improved when more rotation angles are involved in the computation. Specifically, we can see that the F1-Score of using four angles is $2.53$ points higher than that using only one angle. Thus the rotation angle set $\{0, \pi/2, \pi, 3\pi/2\}$ is used in other experiments, unless specified otherwise.

\begin{table}[t]
    \caption{F1-Scores of different rotation angle sets in the FOA module.}
    \label{tab:offset_rcnn_foa}
    \vspace{-2mm}
    \renewcommand{\arraystretch}{1.1}
    \renewcommand{\tabcolsep}{3.0mm}
    \centering
    \begin{tabular}{clc}
        \toprule
        Method                   & Rotation angle set  & F1-Score (\%) \\
        \midrule

        \multirow{4}{*}{LOFT w/ FOA}  
                        & $\{0\}$	                    & 61.78\\
                        & $\{0, \pi/2\}$	            & 63.83\\
                        & $\{0, \pi/2, \pi\}$	        & 63.95\\
                        & $\{0, \pi/2, \pi, 3\pi/2\}$	& \textbf{64.31} \\
        \bottomrule
    \end{tabular}
    \vspace{-3mm}
\end{table}

\textbf{Parameter Sharing in the FOA.} In the FOA module, four convolution (Conv) layers and two fully connected (FC) layers are used in each branch. Thus we implement ablation studies on whether or not to share the parameters of Conv and FC layers. Results are shown in Tab.~\ref{tab:foa_share_parameter}. One can observe that the LOFT achieves the best performance when the parameters of FC layers are shared. Besides, sharing the parameters of FC layers can also reduce the number of model parameters. Therefore, we only share the parameters of FC layers in other experiments.

\begin{table}[t]
    \caption{Results of different network design choices of the FOA. Conv and FC mean the convolution and fully connected layers, respectively.}
    \vspace{-2mm}
    \label{tab:foa_share_parameter}
    \renewcommand{\arraystretch}{1.1}
    \renewcommand{\tabcolsep}{2.0mm}
    \begin{center}
    \begin{tabular}{lccc}
        \toprule
        Method                                  & Share Conv          & Share FC          & F1-Score (\%) \\
        \midrule
        
        \multirow{4}{*}{LOFT w/ FOA}     &  	-	        & -             & 63.84    \\
                                                 & \checkmark	    & -             & 63.22    \\
                                                 & -	            & \checkmark    & \textbf{64.31}   \\
                                                 & \checkmark	    & \checkmark    & 62.94  \\
        \bottomrule
    \end{tabular}
    \end{center}
    \vspace{\fixedvskiptab}
\end{table}

\textbf{Image-level Rotation Augmentation with the FOA.} We compare the performance of image-level rotation augmentation (IRA) with the FOA module. In our experiments of the IRA, the input image and corresponding ground truth are randomly rotated with a rotation angle set $\{0, \pi/2, \pi, 3\pi/2\}$ as the same as in the FOA module. The F1-scores of footprint and roof are both reported in Tab.~\ref{tab:image_level}, where 1x means training with 24 epochs. One we can find that more training time (\#epoch) is needed to obtain more accurate results when the IRA is used. Specifically, the LOFT obtains better footprint F1-Score when the LOFT is trained from 24 epochs to 48 epochs regardless of whether the FOA is used or not ($62.55\%$ to $65.22\%$ with FOA, and $59.88\%$ to $63.75\%$ without FOA). In addition, the FOA module takes only a quarter of the training time (24 epochs) to obtain the comparable result ($64.31\%$ to $64.43\%$) with the IRA trained by 96 epochs. Besides, the FOA module can still improve the accuracy regardless of the training time even with IRA. The other we can see that the trends of F1-Scores of roof and footprint are consistent when the IRA is used since the IRA transforms the roof and footprint simultaneously in the training process. Due to the FOA only rotating the offset feature, it has little effect on the F1-Score of the roof, which also implies the FOA improves the BFE performance by refining the offset prediction.
\begin{table}[t!]
    \caption{Comparison of image-level rotation augmentation (IRA) with the FOA module. 1x means 24 epochs (\%).}
    \label{tab:image_level}
    \renewcommand{\arraystretch}{1.1}
    \renewcommand{\tabcolsep}{1.0mm}
    \vspace{-2mm}
    \centering
    \begin{tabular}{lccccc}
        \toprule
        Method                                  & IRA        & FOA             & $\#$epoch & Footprint F1-Score & Roof F1-Score  \\
        \midrule
        
        \multirow{7}{*}{LOFT}                   &  	-	          & -           & 1x & 61.78 & 67.17    \\
                                                & \checkmark	  & -               & 1x & 59.88 & 66.58   \\
                                                & \checkmark	  & -               & 2x & 63.75 & 68.53    \\
                                                & \checkmark	  & -               & 4x & 64.43 & 68.56    \\
                                                & -	              & \checkmark      & 1x & 64.31 & 67.25 \\
                                                & \checkmark	  & \checkmark      & 1x & 62.55 & 66.71  \\
                                                & \checkmark	  & \checkmark      & 2x & \textbf{65.22} & \textbf{68.71} \\
        \bottomrule
    \end{tabular}
    \vspace{\fixedvskiptab}
\end{table}

\subsection{Discussions}

\textbf{End-point Error of Offset Learning.} In the LOFT, the footprint F1-Score can only indirectly reflect the performance of offset learning. To directly evaluate the performance of offset learning, we compute the object-wise {\em end-point error} (denoted by EPE) in pixels, which is the Euclidean distance between the endpoints of the predicted and ground truth offset vectors. Note that we only calculate the EPE value of the offset vector when its corresponding footprint prediction is true positive. The results of different methods are shown in Tab.~\ref{tab:offset_epe}. One can find that the average EPE of the LOFT is just $5.26$ pixels, and the FOA module can further reduce the error of offset prediction (from $5.26$ to $4.94$), which has the same trend as the footprint F1-Score in Tab.~\ref{tab:influence_foa}.

\begin{table}[t!]
    \caption{The end-point error comparison of LOFT with or without FOA.}
    \label{tab:offset_epe}
    \vspace{-2mm}
    \renewcommand{\arraystretch}{1.2}
    \renewcommand{\tabcolsep}{4.0mm}
    \centering
    \begin{tabular}{ccc}
        \toprule
        Method      & LOFT  & LOFT w/ FOA\\
        \midrule
        Average EPE (pixel) & 5.26  & \textbf{4.94}\\
        \bottomrule
    \end{tabular}
    \vspace{-3mm}
\end{table}

\textbf{Upper Bound Performance.} For offset-based BFE methods, the footprint extraction accuracy largely depends on the prediction accuracies of the roofs and offset vectors. However, if we use the ground truth offsets to replace the predicted offsets, the prediction performances of the roofs and footprints will be the same. Hence, the footprint extraction accuracy is upper bounded by the performance of roof prediction. The performance of our proposed LOFT and Mask R-CNN on the extraction of roofs and footprints is shown in Tab.~\ref{tab:offset_upper}. The performance gap between the roofs and the footprints of our method is just $2.94$ points in F1-Score which is much smaller than the $8.98$ points of Mask R-CNN. Considering that in off-nadir images, the features of building roofs are more notable than building footprints, which implies that our offset-based method is highly effective to BFE problem in off-nadir imagery.

\begin{table}[t]
    \caption{The F1-Scores of roof and footprint extraction by using LOFT or Mask R-CNN.}
    \label{tab:offset_upper}
    \vspace{-2.5mm}
    \renewcommand{\arraystretch}{1.1}
    \renewcommand{\tabcolsep}{3.0mm}
    \centering
    \begin{tabular}{lccc}
        \toprule
        Method                                  & Roof (\%)         & Footprint (\%) & Gap (\%) \\
        \midrule
        LOFT w/ FOA                            & \textbf{67.25}	   & \textbf{64.31} & \textbf{2.94}   \\
        Mask R-CNN~\cite{Mask-R-CNN_2017_ICCV} & 67.10             & 58.12 & 8.98 \\
        \bottomrule
    \end{tabular}
    \vspace{-2mm}
\end{table}

\begin{figure}[t!]
    \centering
    \renewcommand{\tabcolsep}{0.4mm}
    \renewcommand{\arraystretch}{1.0}
    \centering
	\begin{tabular}{ccc}
	\includegraphics[width=0.325\linewidth]{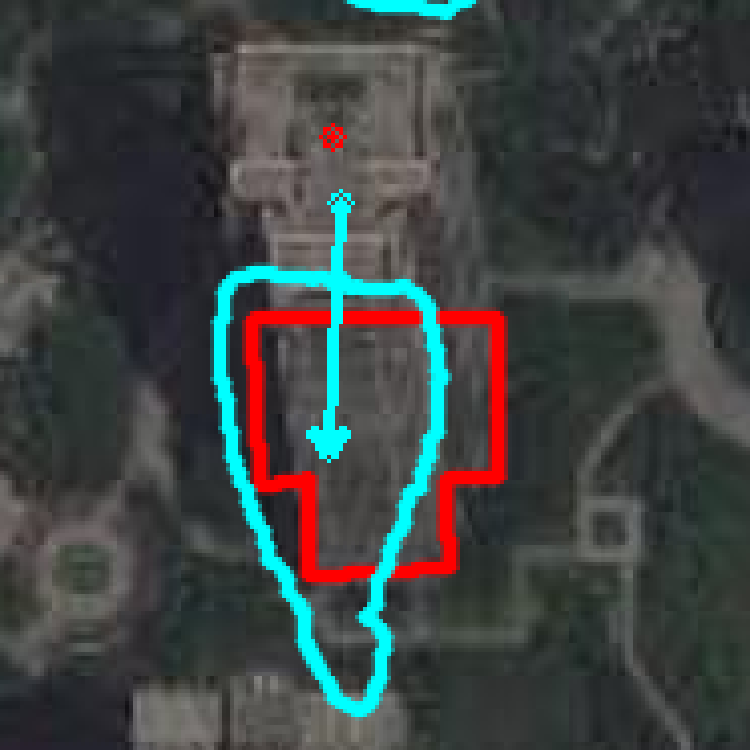} &
	\includegraphics[width=0.325\linewidth]{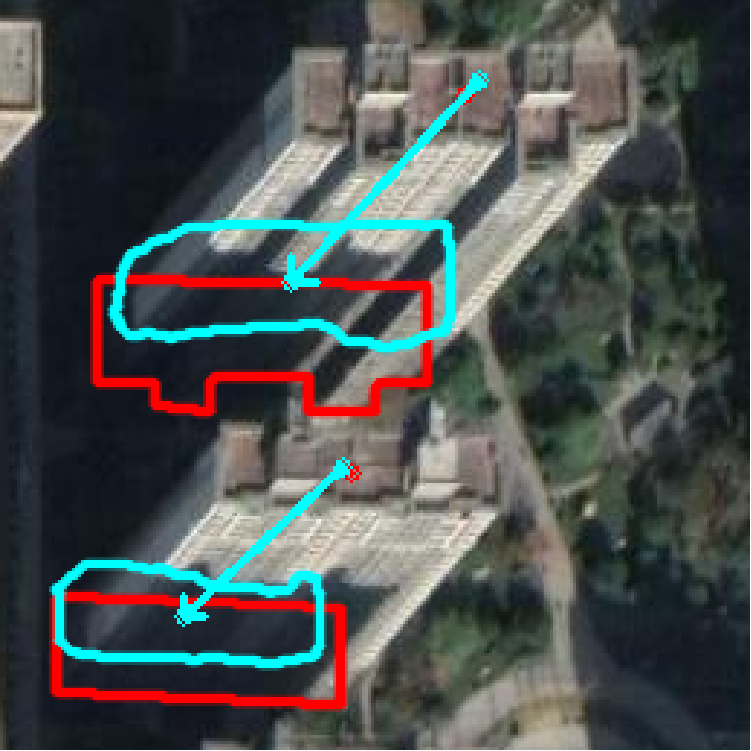} &
    \includegraphics[width=0.325\linewidth]{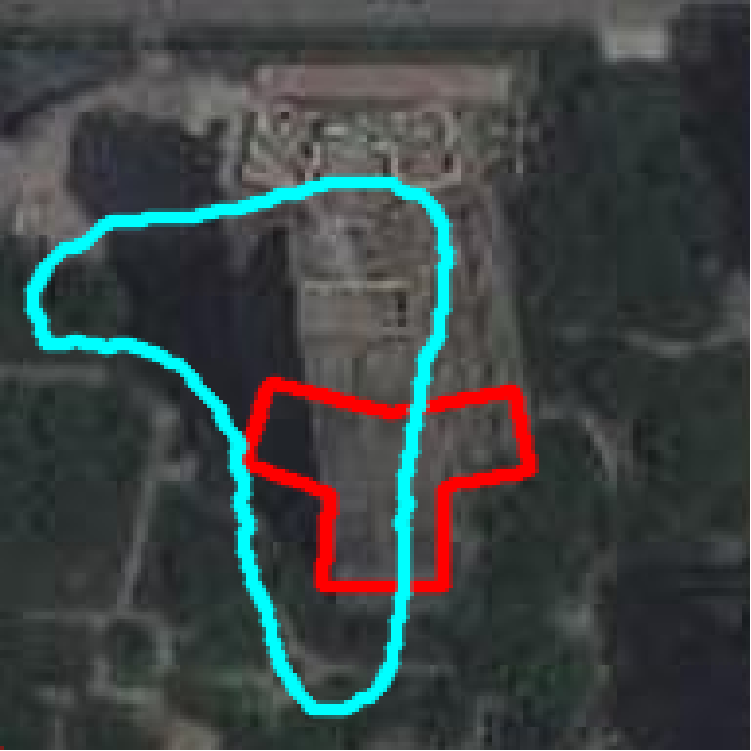}
	\end{tabular}
	
    \vspace{-2mm}
	\caption{Some typical failure cases of LOFT on BONAI dataset. Blue and red mean prediction and ground truth, respectively.}
	\label{fig:failure_cases}
	\vspace{-3mm}
\end{figure}

\textbf{Failure Cases.} Fig.~\ref{fig:failure_cases} reveals three typical failure cases when using the proposed LOFT model, mainly resulting from the prediction errors of building locations, shapes, and offset vectors. 
The failure in the left of Fig.~\ref{fig:failure_cases} occurs as the building roof is confused with the building facade or background, while the middle occurs when the prediction of offset is far from correct. The right one is most challenging as both the shape and the location are difficult to predict. To better handle the BFE problem in off-nadir images and reduce the failure cases, further study may consider facade segmentation as an extra task in a multi-task learning scheme. Besides, roof boundary learning might help to predict more accurate building roofs.

\section{Conclusion}
\label{sec::conclusion}
In this paper, we have addressed the problem of building footprint extraction in off-nadir imagery. To make the most of the property that the building footprint is partially visible while the building roof is fully visible in off-nadir images, we propose the LOFT scheme to decouple the BFE problem to the building roof extraction superimposing a roof-to-footprint offset vector regression. The proposed {offset head} can be easily applied to any top-down instance segmentation method. Moreover, a simple but effective feature-level offset augmentation module is proposed to refine the offset vector prediction further, avoiding significant extra computation in traditional image-based augmentation. A new dataset, \ie, BONAI, is also created to train and evaluate BFE models for Off-nadir aerial images. Experimental results on BONAI demonstrate the superiority of our method.

\ifCLASSOPTIONcaptionsoff
  \newpage
\fi

\bibliographystyle{IEEEtran}
\bibliography{tpami-name,main}

\begin{thebibliography}{10}
\providecommand{\url}[1]{#1}
\csname url@samestyle\endcsname
\providecommand{\newblock}{\relax}
\providecommand{\bibinfo}[2]{#2}
\providecommand{\BIBentrySTDinterwordspacing}{\spaceskip=0pt\relax}
\providecommand{\BIBentryALTinterwordstretchfactor}{4}
\providecommand{\BIBentryALTinterwordspacing}{\spaceskip=\fontdimen2\font plus
\BIBentryALTinterwordstretchfactor\fontdimen3\font minus
  \fontdimen4\font\relax}
\providecommand{\BIBforeignlanguage}[2]{{%
\expandafter\ifx\csname l@#1\endcsname\relax
\typeout{** WARNING: IEEEtran.bst: No hyphenation pattern has been}%
\typeout{** loaded for the language `#1'. Using the pattern for}%
\typeout{** the default language instead.}%
\else
\language=\csname l@#1\endcsname
\fi
#2}}
\providecommand{\BIBdecl}{\relax}
\BIBdecl

\bibitem{Birth_Death-2011-TPAMI}
C.~Benedek, X.~Descombes, and J.~Zerubia, ``Building development monitoring in
  multitemporal remotely sensed image pairs with stochastic birth-death
  dynamics,'' \emph{{TPAMI}}, vol.~34, no.~1, pp. 33--50, 2011.

\bibitem{SDF_2020_CVPR}
J.~Mahmud, T.~Price, A.~Bapat, and J.-M. Frahm, ``Boundary-aware 3d building
  reconstruction from a single overhead image,'' in \emph{{CVPR}}, 2020, pp.
  441--451.

\bibitem{Signed_distance_function_2017_TPAMI}
J.~Yuan, ``Learning building extraction in aerial scenes via convolutional
  network,'' \emph{{TPAMI}}, vol.~40, no.~11, pp. 2793--2798, 2017.

\bibitem{DSAC_CVPR_2018}
D.~Marcos, D.~Tuia, B.~Kellenberger, L.~Zhang, M.~Bai, R.~Liao, and R.~Urtasun,
  ``Learning deep structured active contours end-to-end,'' in \emph{{CVPR}},
  2018, pp. 8877--8885.

\bibitem{Conv_MPN_CVPR_2020}
F.~Zhang, N.~Nauata, and Y.~Furukawa, ``Conv-mpn: Convolutional message passing
  neural network for structured outdoor architecture reconstruction,'' in
  \emph{{CVPR}}, 2020, pp. 2798--2807.

\bibitem{BFE3_1999_TPAMI}
J.~A. Shufelt, ``Performance evaluation and analysis of monocular building
  extraction from aerial imagery,'' \emph{{TPAMI}}, vol.~21, no.~4, pp.
  311--326, 1999.

\bibitem{BFE1_2007_TPAMI}
M.~Ortner, X.~Descombes, and J.~Zerubia, ``A marked point process of rectangles
  and segments for automatic analysis of digital elevation models,''
  \emph{{TPAMI}}, vol.~30, no.~1, pp. 105--119, 2007.

\bibitem{svm_building_2007_ISPRSJ}
J.~Inglada, ``Automatic recognition of man-made objects in high resolution
  optical remote sensing images by svm classification of geometric image
  features,'' \emph{{ISPRS J. Photogramm. Remote Sens.}}, vol.~62, no.~3, pp.
  236--248, 2007.

\bibitem{BFE2_2008_TPAMI}
F.~Lafarge, X.~Descombes, J.~Zerubia, and M.~Pierrot-Deseilligny, ``Structural
  approach for building reconstruction from a single dsm,'' \emph{{TPAMI}},
  vol.~32, no.~1, pp. 135--147, 2008.

\bibitem{Darnet_2019_CVPR}
D.~Cheng, R.~Liao, S.~Fidler, and R.~Urtasun, ``Darnet: Deep active ray network
  for building segmentation,'' in \emph{{CVPR}}, 2019, pp. 7431--7439.

\bibitem{MAP-Net_2020_TGRS}
Q.~Zhu, C.~Liao, H.~Hu, X.~Mei, and H.~Li, ``Map-net: Multiple attending path
  neural network for building footprint extraction from remote sensed
  imagery,'' \emph{{IEEE Trans. Geosci. Remote Sensing}}, pp. 1--13, 2020.

\bibitem{Polygon_Building_2020_CVPR}
M.~Li, F.~Lafarge, and R.~Marlet, ``Approximating shapes in images with
  low-complexity polygons,'' in \emph{{CVPR}}, 2020, pp. 8633--8641.

\bibitem{Learning_Geocentric_Object_Pose_2020_CVPR}
G.~Christie, R.~R. R.~M. Abujder, K.~Foster, S.~Hagstrom, G.~D. Hager, and
  M.~Z. Brown, ``Learning geocentric object pose in oblique monocular images,''
  in \emph{{CVPR}}, 2020, pp. 14\,512--14\,520.

\bibitem{Mask-R-CNN_2017_ICCV}
K.~He, G.~Gkioxari, P.~Dollar, and R.~Girshick, ``{Mask R-CNN},'' in
  \emph{{ICCV}}, 2017, pp. 2961--2969.

\bibitem{INRIA_2017_IGARSS}
E.~Maggiori, Y.~Tarabalka, G.~Charpiat, and P.~Alliez, ``Can semantic labeling
  methods generalize to any city? the inria aerial image labeling benchmark,''
  in \emph{{Proc. Int. Geosci. Remote Sensing Symposium}}.\hskip 1em plus 0.5em
  minus 0.4em\relax IEEE, 2017.

\bibitem{ISPRS_2018}
``{ISPRS} 2d semantic labeling contest,''
  \url{http://www2.isprs.org/commissions/comm3/wg4/semantic-labeling.html},
  2018.

\bibitem{DSTL_2018_kaggle}
``{DSTL}-kaggle,'' \url{http://www.kaggle.com/c/dstl- satellite- imagery-
  feature- detection}, 2018.

\bibitem{WHU_2018_TGRS}
S.~Ji, S.~Wei, and M.~Lu, ``Fully convolutional networks for multisource
  building extraction from an open aerial and satellite imagery data set,''
  \emph{{IEEE Trans. Geosci. Remote Sensing}}, vol.~57, no.~1, pp. 574--586,
  2019.

\bibitem{SpaceNet_MVOI_2019_CVPR}
N.~Weir, D.~Lindenbaum, A.~Bastidas, A.~V. Etten, S.~McPherson, J.~Sherm,
  V.~Kumar, and H.~Tang, ``Spacenet mvoi: a multi-view overhead imagery
  dataset,'' in \emph{{CVPR}}, 2019, pp. 992--1001.

\bibitem{Synthinel-1_2020_WACV}
F.~Kong, B.~Huang, K.~Bradbury, and J.~Malof, ``The synthinel-1 dataset: a
  collection of high resolution synthetic overhead imagery for building
  segmentation,'' in \emph{{WACV}}, 2020, pp. 1814--1823.

\bibitem{PANet_2018_CVPR}
S.~Liu, L.~Qi, H.~Qin, J.~Shi, and J.~Jia, ``Path aggregation network for
  instance segmentation,'' in \emph{{CVPR}}, 2018, pp. 8759--8768.

\bibitem{HTC_2019_CVPR}
K.~Chen, J.~Pang, J.~Wang, Y.~Xiong, X.~Li, S.~Sun, W.~Feng, Z.~Liu, J.~Shi,
  W.~Ouyang \emph{et~al.}, ``Hybrid task cascade for instance segmentation,''
  in \emph{{CVPR}}, 2019, pp. 4974--4983.

\bibitem{HRNet_2019_arXiv}
K.~Sun, Y.~Zhao, B.~Jiang, T.~Cheng, B.~Xiao, D.~Liu, Y.~Mu, X.~Wang, W.~Liu,
  and J.~Wang, ``High-resolution representations for labeling pixels and
  regions,'' \emph{CoRR}, vol. abs/1904.04514, 2019.

\bibitem{Faster-R-CNN_2015_NIPS}
S.~Ren, K.~He, R.~Girshick, and J.~Sun, ``{Faster R-CNN}: Towards real-time
  object detection with region proposal networks,'' in \emph{{NeurIPS}}, 2015,
  pp. 91--99.

\bibitem{Topological_Structural_Analysis_1985_CVGIP}
S.~Suzuki \emph{et~al.}, ``Topological structural analysis of digitized binary
  images by border following,'' \emph{Computer Vision, Graphics, and Image
  Processing}, vol.~30, no.~1, pp. 32--46, 1985.

\bibitem{Featmatch_2020_ECCV}
C.-W. Kuo, C.-Y. Ma, J.-B. Huang, and Z.~Kira, ``Featmatch: Feature-based
  augmentation for semi-supervised learning,'' in \emph{{ECCV}}.\hskip 1em plus
  0.5em minus 0.4em\relax Springer, 2020, pp. 479--495.

\bibitem{STN_2015_NIPS}
M.~Jaderberg, K.~Simonyan, A.~Zisserman \emph{et~al.}, ``Spatial transformer
  networks,'' in \emph{{NeurIPS}}, 2015, pp. 2017--2025.

\bibitem{DOTA_2018_CVPR}
G.-S. Xia, X.~Bai, J.~Ding, Z.~Zhu, S.~Belongie, J.~Luo, M.~Datcu, M.~Pelillo,
  and L.~Zhang, ``{DOTA}: A large-scale dataset for object detection in aerial
  images,'' in \emph{{CVPR}}, 2018, pp. 3974--3983.

\bibitem{DOTA_PAMI}
J.~Ding, N.~Xue, G.-S. Xia, X.~Bai, W.~Yang, M.~Yang, S.~Belongie, J.~Luo,
  M.~Datcu, M.~Pelillo, and L.~Zhang, ``Object detection in aerial images: A
  large-scale benchmark and challenges,'' \emph{TPAMI}, pp. 1--1, 2021.

\bibitem{ResNet_2016_CVPR}
K.~He, X.~Zhang, S.~Ren, and J.~Sun, ``Deep residual learning for image
  recognition,'' in \emph{{CVPR}}, 2016, pp. 770--778.

\bibitem{FPN_2017_CVPR}
T.-Y. Lin, P.~Dollar, R.~Girshick, K.~He, B.~Hariharan, and S.~Belongie,
  ``Feature pyramid networks for object detection,'' in \emph{{CVPR}}, 2017,
  pp. 2117--2125.

\bibitem{Boundary_IoU_2021_CVPR}
B.~Cheng, R.~Girshick, P.~Dollar, A.~C. Berg, and A.~Kirillov, ``Boundary iou:
  Improving object-centric image segmentation evaluation,'' in \emph{{CVPR}},
  2021, pp. 15\,334--15\,342.

\end{thebibliography}

\end{document}